\documentclass{article}
\usepackage[T1]{fontenc}
\usepackage{iclr2027_conference,times}

\usepackage{amsmath,amsfonts,bm}

\def\Figref#1{Figure~\ref{#1}}

\def\Secref#1{Section~\ref{#1}}

\def\eqref#1{equation~\ref{#1}}
\def\Eqref#1{Equation~\ref{#1}}

\def\1{\bm{1}}

\def\vc{{\bm{c}}}

\def\vf{{\bm{f}}}
\def\vg{{\bm{g}}}
\def\vh{{\bm{h}}}

\def\vr{{\bm{r}}}

\def\vu{{\bm{u}}}
\def\vv{{\bm{v}}}

\def\vx{{\bm{x}}}

\DeclareMathAlphabet{\mathsfit}{\encodingdefault}{\sfdefault}{m}{sl}
\SetMathAlphabet{\mathsfit}{bold}{\encodingdefault}{\sfdefault}{bx}{n}

\usepackage{hyperref}
\hypersetup{hidelinks}
\usepackage{url}
\usepackage{amsmath,amssymb,bm}
\usepackage{booktabs,multirow,array}
\usepackage{graphicx}
\usepackage{flafter}
\usepackage{xcolor}
\usepackage{caption,subcaption}
\usepackage{tikz}
\usetikzlibrary{arrows.meta,calc,positioning}

\newcommand{\method}{PTD}
\newcommand{\teacher}{u_{\mathrm{T}}}
\newcommand{\sg}{\mathrm{sg}}

\title{Parametric Trajectory Distillation for\\ Few-Step Video Generation}

\author{Lan Feng$^{1,2}$, Peter Karkus$^{2}$, Maximilian Igl$^{2}$, Julius Berner$^{2}$, Yuxiao Chen$^{2}$, \\
\textbf{Shuhan Tan$^{2}$, Alexandre Alahi$^{1}$, Boris Ivanovic$^{2}$, Marco Pavone$^{2,3}$} \\
$^{1}$EPFL \quad $^{2}$NVIDIA \quad $^{3}$Stanford University}
\iclrfinalcopy
\hypersetup{pdftitle={Parametric Trajectory Distillation for Few-Step Video Generation},
  pdfauthor={Lan Feng, Peter Karkus, Maximilian Igl, Julius Berner, Yuxiao Chen, Shuhan Tan, Alexandre Alahi, Boris Ivanovic, Marco Pavone}}

\begin{document}

\maketitle
\lhead{Preprint}\vspace{-0.1in}

\noindent
\begin{minipage}{\linewidth}
  \centering
  \includegraphics[width=\linewidth]{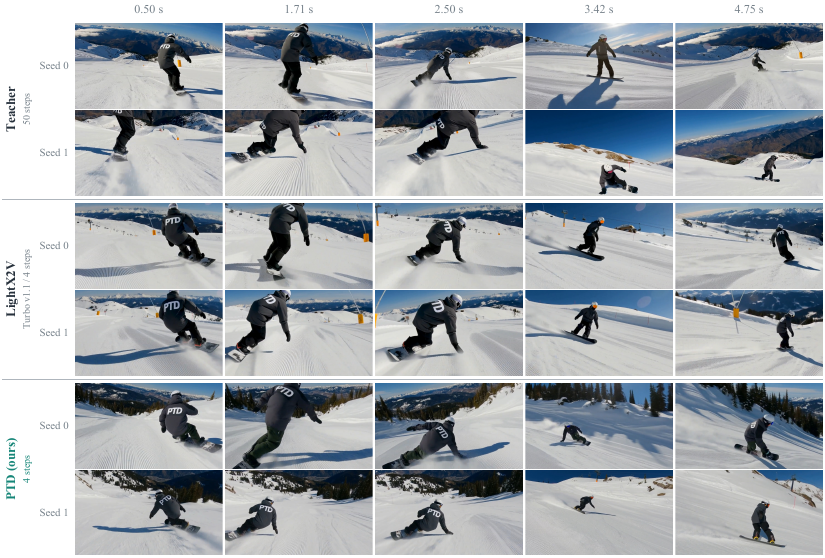}
  \captionsetup{font=footnotesize,skip=3pt}
  \captionof{figure}{\textbf{Natural motion and seed diversity in four steps (MiniMax-H3).} Like the teacher, \method{} carves deep turns with seeds that differ in framing and background, whereas the two LightX2V Turbo (DMD) seeds nearly repeat framing and pose. Over 100 prompts, \method{} raises GPT-6 Astra diversity from 1.55 to 2.81 (Section~\ref{sec:h3}). Each row: one seed at five matching timestamps. Teacher: 50 steps; students: four. Prompt: Appendix~\ref{app:teaser-snowboard}.}
  \label{fig:teaser}
\end{minipage}
\par\vspace{4pt}

\begin{abstract}
Video diffusion and flow models require many sequential evaluations, making generation computationally expensive.
Few-step distillation reduces this cost but poses a capacity allocation problem: a student must match the teacher's iterative generation with far less sequential computation.
Existing trajectory methods ask the student to reproduce teacher transitions that are highly curved at high noise, which can exceed its capacity and degrade fine detail.
We introduce Parametric Trajectory Distillation (\method), which lets the student parameterize teacher trajectory segments as polynomials and learn from teacher guidance along its own predicted path.
\method{} is designed to let the learned curvature adapt to the backbone's predictive capacity, preserving motion and diversity.
The curvature head is used only in training; inference keeps the original backbone architecture.
On Wan2.1-14B, four-step \method{} sets a new state of the art for trajectory distillation, significantly improving dynamic quality and naturalness over PDD, the best-performing trajectory-only method on this model, under the same training setting.
On the 33B audio-video MiniMax-H3, LoRA-trained \method{} significantly improves diversity and naturalness over the state-of-the-art LightX2V Turbo.
Blinded human votes give \method{} 55.1\% and 63.4\% preference shares against PDD and LightX2V Turbo. Project page: \href{https://alan-lanfeng.github.io/PTD/}{alan-lanfeng.github.io/PTD}.
\end{abstract}

\section{Introduction}
\label{sec:intro}

Video diffusion and flow models generate samples through many sequential network evaluations~\citep{wan2025,lipman2023flow,liu2023rectified}.
Few-step distillation compresses this iterative process into a handful of student evaluations, often retaining the teacher's backbone.
The student may therefore have the same parameter count but far less sequential computation with which to reproduce the teacher's generation process.
This makes distillation a problem of capacity allocation: supervision must help the student use its limited computation to approximate a much longer generation process at comparable sample quality.

Existing approaches address this challenge through trajectory supervision or distribution matching.
Trajectory objectives~\citep{song2023consistency,lu2025scm} train the student to model the teacher's full noise-to-sample trajectory, preserving motion and diversity.
This objective is mode-covering: when a few-step student lacks the capacity to represent complex teacher transitions, regression pulls its predictions toward an average of the targets, blurring fine detail~\citep{zheng2025rcm}.
Recently, Parallel Decoding Distillation (PDD) achieved strong visual detail with trajectory supervision alone by regressing parallel interval-velocity outputs to teacher Runge--Kutta targets~\citep{anonymous2026pdd}.
Distribution Matching Distillation (DMD) avoids this averaging: its mode-seeking objective evaluates the teacher only on the student's own samples, recovering sharp detail at the cost of substantially reduced sample diversity~\citep{yin2024dmd,zheng2025rcm}.
Hybrid methods such as rCM and AnyFlow combine the two paradigms~\citep{zheng2025rcm,gu2026anyflow}, but the added objectives increase training cost and complexity.
Rather than stacking objectives, is there a more elegant way to combine the strengths of both paradigms?

\begin{figure}[t]
  \centering
  \resizebox{0.9\linewidth}{!}{\input{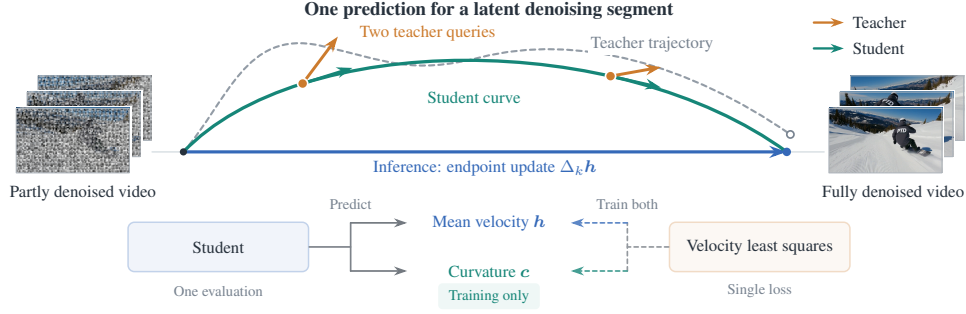}}
  \caption{\textbf{Parametric Trajectory Distillation.}
  One student evaluation predicts mean velocity $\vh$ and curve coefficients $\vc$ for a latent denoising segment.
  Teacher velocities queried on the student curve train both outputs with one least-squares loss.
  Inference uses only $\vh$ for the endpoint update; the curvature head is removed.
  Illustrative video thumbnails depict a segment ending at the data endpoint; the same construction applies between intermediate noise levels.
  Arrows show step-scaled velocities; two queries are the default.}
  \label{fig:overview}
\end{figure}

We introduce \emph{Parametric Trajectory Distillation} (\method) to align trajectory learning with the student's predictive capacity.
A single student evaluation predicts a polynomial approximation of a teacher trajectory segment, and teacher velocities are queried along this predicted path (\Figref{fig:overview}).
Like distribution matching, which evaluates the teacher only on the student's own samples, \method{} thus supervises the student on the path it currently predicts, while keeping the noise-to-sample correspondence of trajectory supervision.
This design is built to retain the teacher's motion and diversity without demanding trajectory detail that the student's backbone cannot reliably predict. %
The curve serves as a training scaffold: its mean velocity directly supplies the endpoint update, while the curvature head is discarded at inference, preserving the original inference architecture.
Both predictions are trained with a single velocity-regression loss, without computing Jacobian-vector products or training an auxiliary critic.
These properties make \method{} a simple recipe for scaling trajectory distillation to large backbones, including through parameter-efficient LoRA fine-tuning.

We distill MiniMax-H3~\citep{minimaxh3}, a 33B joint audio-video model, into a four-step generator using LoRA and trajectory supervision alone.
Against LightX2V Turbo~\citep{modeltc2026h3turbo}, a state-of-the-art four-step DMD method on the H3 Acceleration Arena~\citep{h3arena2026}, \method{} significantly improves diversity (2.81 vs.\ 1.55) and naturalness as measured by GPT-6 Astra, while matching its dynamic and static quality (Figures~\ref{fig:teaser} and~\ref{fig:astra_main}).
We also apply \method{} with full fine-tuning to Wan2.1-14B~\citep{wan2025}, a 14B text-to-video model.
Under the same training setting, it outperforms PDD~\citep{anonymous2026pdd}, the best-performing trajectory-only method reported on this model, significantly improving dynamic quality and naturalness under the same judge and achieving a higher overall VBench score~\citep{huang2024vbench}.
Blinded human votes give \method{} 63.4\% and 55.1\% preference shares against LightX2V Turbo and PDD.

Our contributions are:
(i) a parametric trajectory distillation formulation that queries teacher velocities along student-predicted polynomial paths and retains only the predicted mean velocity at inference;
(ii) analyses showing that teacher guidance along the student's own path preserves fine detail and that the student learns a shrunken, learnable version of the teacher's curvature;
and (iii) state-of-the-art four-step video distillation with trajectory supervision alone, improving over LightX2V Turbo on the 33B MiniMax-H3 and surpassing PDD on Wan2.1-14B.

\section{Related work}
\label{sec:related}

\paragraph{Endpoint and flow-map learning.}
Consistency models learn flow maps~\citep{song2023consistency,kim2024ctm}, whose semigroup property underlies progressive distillation and shortcut models~\citep{salimans2022progressive,frans2025shortcut,boffi2025howto}.
MeanFlow and its follow-ups learn interval-average velocity~\citep{geng2025meanflow,geng2026imf,guo2025splitmeanflow}.
\method{}'s $\vh$ is this mean velocity, fit to teacher velocities rather than to self-consistency targets.

\paragraph{Explicit trajectories and local supervision.}
These methods differ in where the teacher is queried.
DSNO learns continuous trajectories from teacher-generated states~\citep{zheng2023dsno}; PDD, the best-performing trajectory-only method on Wan2.1-14B, regresses fusible interval-velocity outputs to teacher Runge--Kutta targets~\citep{anonymous2026pdd}.
Flow Map Matching and Physics Informed Distillation query the teacher at student-predicted states~\citep{boffi2024flowmap,tee2024pid}; $\pi$-Flow imitates teacher velocities along rollouts of a compact policy integrated at inference~\citep{chen2026piflow}.
\method{} queries the teacher along its own predicted curve, takes velocities from analytic basis derivatives instead of differentiating the student in time, and drops the curvature head at inference.

\paragraph{Distribution matching for video.}
DMD and DMD2 match generated distributions through score-based supervision~\citep{yin2024dmd,yin2024dmd2}; for video, rCM regularizes consistency with score supervision~\citep{zheng2025rcm}, and AnyFlow adds on-policy distribution refinement~\citep{gu2026anyflow}.
\method{} uses trajectory supervision alone, without any distribution-level loss, and still significantly improves Astra diversity and naturalness over a state-of-the-art DMD student (\Secref{sec:h3}).
Appendix~\ref{app:related} expands these comparisons.

\section{Parametric Trajectory Distillation}
\label{sec:method}

\method{} predicts a denoising segment in one student evaluation and queries the teacher along that predicted path (\Figref{fig:overview}).
A single velocity-regression objective trains both its displacement and interior shape; inference retains only the endpoint update.

\paragraph{Setup.}
\label{sec:setup}
Let $\teacher(\vx,t\mid y)$ be a pretrained flow model conditioned on a text prompt $y$~\citep{lipman2023flow,liu2023rectified}.
The teacher generates samples by integrating $\mathrm d\vx/\mathrm dt=\teacher(\vx,t\mid y)$ from noise at $t=1$ to data at $t=0$.
We distill this process onto a fixed decreasing grid $t_0>\cdots>t_N$, with $\Delta_k=t_{k+1}-t_k<0$.
We refer to each interval, which one student evaluation traverses, as a \emph{hop}. For hop $k$, starting at state $\vx_k$ at time $t_k$, define normalized progress $\tau\in[0,1]$ and physical time $s(\tau)=t_k+\tau\Delta_k$.
We suppress the conditioning where clear.

\subsection{One prediction for a trajectory segment}
\label{sec:path}
A single forward pass of the student with parameters $\theta$ at $(\vx_k,t_k)$ predicts a mean-velocity vector $\vh_\theta\in\mathbb R^D$ and $K$ curvature coefficient vectors $\vc_\theta=(\vc_{\theta,0},\ldots,\vc_{\theta,K-1})$ with $\vc_{\theta,j}\in\mathbb R^D$, where $D$ is the latent dimension.
These outputs define the entire segment:
\begin{equation}
  \vx_{\theta,k}(\tau)=\vx_k+\Delta_k\left[\tau\vh_\theta+\sum_{j=0}^{K-1}b_j(\tau)\vc_{\theta,j}\right],
  \qquad b_j(\tau)=\tau(\tau-1)P_j(2\tau-1).
  \label{eq:path}
\end{equation}
The \emph{curvature basis functions} $b_j$ are built from degree-$j$ Legendre polynomials $P_j$ and vanish at both endpoints: $b_j(0)=b_j(1)=0$.
Thus the path connects $\vx_k$ to $\vx_k+\Delta_k\vh_\theta$ for any $\vc_\theta$.
The curvature output $\vc_\theta$ controls only the interior shape, with $\vc_\theta=0$ recovering the straight chord.

\paragraph{Mean velocity and local velocity.}
Advancing a hop in one step requires its mean velocity, while teacher queries provide local velocities.
We relate these quantities by differentiating the curve with respect to physical time, holding the network outputs fixed:
\begin{equation}
  \vv_{\theta,k}(\tau)=\frac{\mathrm d\vx_{\theta,k}}{\mathrm ds}
  =\vh_\theta+\sum_{j=0}^{K-1}d_j(\tau)\vc_{\theta,j},
  \qquad d_j(\tau)=\frac{\mathrm d b_j(\tau)}{\mathrm d\tau}.
  \label{eq:tangent}
\end{equation}
Only the known basis functions are differentiated, so additional curve positions and velocities require no further student forward pass or network Jacobian-vector product.
The endpoint constraints imply $\int_0^1d_j(\tau)\,\mathrm d\tau=0$, and hence
\begin{equation}
  \int_0^1\vv_{\theta,k}(\tau)\,\mathrm d\tau=\vh_\theta,
  \qquad \vx_{\theta,k}(1)=\vx_k+\Delta_k\vh_\theta.
  \label{eq:endpoint}
\end{equation}
Thus $\vh_\theta$ is exactly the mean velocity of the represented student curve, for any network weights.

\subsection{A single loss on the student curve}
\label{sec:target}
To train the segment, we sample $m$ positions $\tau_i\in(0,1)$ and query the teacher at the corresponding points on the current student curve:
\begin{equation}
  \vx_i=\sg\!\left[\vx_{\theta,k}(\tau_i)\right],\qquad
  s_i=s(\tau_i),\qquad
  \vu_i=\sg\!\left[\teacher(\vx_i,s_i\mid y)\right],
  \label{eq:reads}
\end{equation}
where $\sg$ denotes stop-gradient.
We fit the predicted velocity to these teacher velocities with one least-squares loss:
\begin{equation}
  \mathcal L_{\mathrm{PTD}}=\sum_{i=1}^{m}\omega_i
  \left\|\vh_\theta+\sum_{j=0}^{K-1}d_j(\tau_i)\vc_{\theta,j}-\vu_i\right\|^2,
  \qquad \omega_i>0,\quad\sum_i\omega_i=1.
  \label{eq:ls}
\end{equation}
During each update, gradients flow through both outputs in the predicted velocity; the teacher targets and their query states are detached.
The queries are recomputed from the student's predictions on subsequent updates.
\Eqref{eq:ls} estimates the Lagrangian map distillation loss~\citep{boffi2024flowmap,boffi2025howto} with grid start times and polynomial maps; Appendix~\ref{app:relations} relates it to TVM, $\pi$-Flow, and PDD.

\paragraph{Joint supervision of displacement and shape.}
Each local velocity residual supervises $\vh_\theta$ directly and the curve coefficients through the weights $d_j(\tau_i)$.
For the uniformly integrated objective, the zero-integral property of $d_j$ makes the expected output gradient train $\vh_\theta$ toward the teacher velocity averaged along the current student curve.
The same residual trains $\vc_\theta$ to capture the teacher's velocity variation along that path.
No separate mean-velocity or curvature target is required.
Appendix~\ref{app:basis} derives these output gradients and distinguishes the mean along the student curve from the mean along the true teacher trajectory.
Under a Lipschitz condition on the teacher field, the integrated residual bounds the endpoint error; exact agreement throughout the interval recovers the teacher trajectory (Appendix~\ref{app:family}).

\begin{figure}[!t]
  \centering
  \resizebox{0.9\linewidth}{!}{\begingroup%
\definecolor{ptdInk}{HTML}{27313C}%
\definecolor{ptdMuted}{HTML}{788390}%
\definecolor{ptdBlue}{HTML}{3569B8}%
\definecolor{ptdTeal}{HTML}{168678}%
\definecolor{ptdOrange}{HTML}{CC7B2D}%
\definecolor{ptdRule}{HTML}{DEE4E8}%
\begin{tikzpicture}[
  x=1cm,y=1cm,
  >={Stealth[length=1.5mm,width=1.1mm]},
  font=\rmfamily\fontsize{8}{9.5}\selectfont,
  text=ptdInk,line cap=round,line join=round,
  panel/.style={anchor=west,inner sep=0pt,font=\rmfamily\bfseries\fontsize{9}{10.5}\selectfont},
  label/.style={inner sep=1.5pt},
  teacher/.style={draw=ptdMuted,line width=1.0pt,dash pattern=on 2.2pt off 1.5pt},
  student/.style={draw=ptdTeal,line width=1.35pt},
  note/.style={rounded corners=2pt,minimum height=.43cm,inner xsep=5pt,inner ysep=2pt}
]
  \path[use as bounding box] (0,0) rectangle (14.15,4.56);
  \def\referencepath{
    (.28,1.40)
    .. controls (.55,2.05) and (.55,3.52) .. (1.15,3.36)
    .. controls (1.80,3.18) and (1.21,2.12) .. (1.82,2.14)
    .. controls (2.40,2.15) and (2.24,3.34) .. (2.90,3.10)
    .. controls (3.60,2.85) and (3.58,2.24) .. (4.24,2.43)
    .. controls (4.88,2.61) and (5.46,1.83) .. (6.45,1.87)
  }
  \foreach \shift in {0,7.40} {
    \begin{scope}[xshift=\shift cm]
      \node[text=ptdMuted,font=\rmfamily\fontsize{7.4}{8.5}\selectfont] at (1.53,3.65) {High-noise portion};
    \end{scope}
  }

  \node[panel] at (0,4.37) {(a) Teacher-marched target};
  \draw[teacher] \referencepath;
  \foreach \p in {(1.15,3.36),(1.82,2.14),(2.90,3.10),(4.24,2.43)}
    \fill[ptdMuted] \p circle (1.6pt);
  \draw[->,ptdBlue,line width=1.1pt] (.28,1.40) -- (6.45,1.87);
  \fill[ptdInk] (.28,1.40) circle (1.7pt);
  \draw[ptdBlue,fill=white,line width=1.0pt] (6.45,1.87) circle (2.2pt);
  \node[label,text=ptdMuted] at (4.75,3.21) {Teacher trajectory};
  \node[label,text=ptdBlue] at (4.00,1.28) {Predict the endpoint};
  \node[label,text=ptdBlue,anchor=east] at (6.72,2.21) {Target};
  \node[note,draw=ptdRule,fill=ptdBlue!4] (capacity) at (1.59,.59) {Limited student capacity};
  \draw[->,ptdMuted,line width=.7pt] (3.23,.59) -- (3.64,.59);
  \node[label,text=ptdOrange,anchor=west] at (3.81,.59) {Risk of blurred detail};

  \begin{scope}[xshift=7.40cm]
    \node[panel] at (0,4.37) {(b) Student-curve supervision};
    \draw[teacher,draw=ptdMuted!45] \referencepath;
    \draw[ptdMuted!55,fill=white,line width=.8pt] (6.45,1.87) circle (1.8pt);
    \draw[student] (.28,1.40)
      .. controls (1.45,2.78) and (4.40,2.39) .. (6.45,1.54)
      coordinate[pos=.22] (q1) coordinate[pos=.64] (q2);
    \fill[ptdInk] (.28,1.40) circle (1.7pt);
    \fill[ptdTeal] (6.45,1.54) circle (2.0pt);
    \draw[->,ptdOrange,line width=1.15pt] (q1) -- ++(.40,.68);
    \draw[->,ptdOrange,line width=1.15pt] (q2) -- ++(.62,.15);
    \foreach \p in {q1,q2} {
      \fill[white] (\p) circle (2.8pt);
      \fill[ptdOrange] (\p) circle (2.0pt);
    }
    \node[label,text=ptdMuted!85] at (4.80,3.21) {Same teacher reference};
    \node[label,text=ptdTeal] at (3.90,1.49) {Student-predicted path};
    \node[note,draw=ptdTeal!20,fill=ptdTeal!5,text=ptdTeal] at (1.56,.59) {Learned approximation};
    \draw[->,ptdMuted,line width=.7pt] (3.12,.59) -- (3.56,.59);
    \node[label,text=ptdOrange,anchor=west] at (3.71,.59) {Capacity-aligned guidance};
  \end{scope}
  \node[text=ptdMuted,font=\rmfamily\fontsize{7}{8}\selectfont] at (7.07,.03) {Schematic of the capacity-allocation rationale};
\end{tikzpicture}%
\endgroup%
}
  \caption{\textbf{Where should the student receive teacher supervision?}
  \textbf{(a)} Dense teacher integration constructs an endpoint target at the cost of many sequential evaluations; on strongly curved, high-noise hops, a limited-capacity student cannot learn this transition well, and fine detail softens.
  \textbf{(b)} \method{} queries teacher velocities on the student's current approximation, jointly training its displacement and shape through local guidance.
  Section~\ref{sec:ablations} compares these supervision choices.}
  \label{fig:student_path_supervision}
\end{figure}
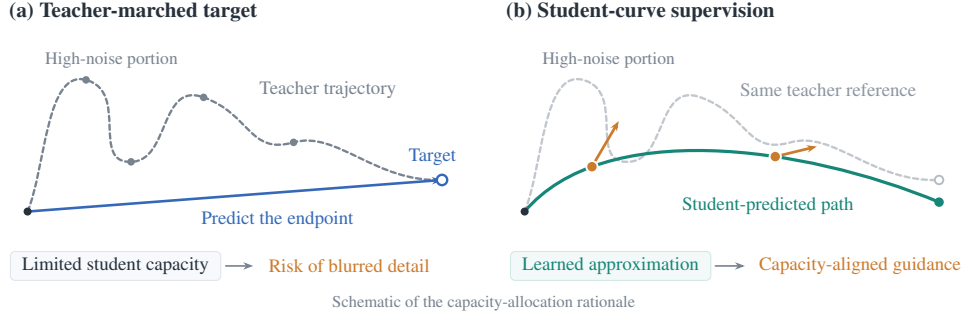

\paragraph{Why query the student curve?}
A straightforward way to supervise $\vh_\theta$ is to integrate the teacher ODE densely over a hop and regress to the resulting displacement divided by $\Delta_k$.
Each such target costs many sequential teacher evaluations and is hardest to learn on high-noise hops, where the teacher trajectory curves most strongly: on H3 teacher trajectories, two curve terms capture only $46\%$ of the within-hop velocity variation near the noise end, versus $90\%$ on the final hop (Figure~\ref{fig:curve_probe}).
Marching every hop softens fine detail (Section~\ref{sec:ablations}).
\method{} instead queries teacher velocities along the curve predicted by the student (\Figref{fig:student_path_supervision}); these local targets avoid dense integration and co-evolve with the predicted coefficients, so supervision always lies on the path the student currently represents.
The final hop is much smoother, so H3 marches only that hop; we empirically found that this improves detail at about $20\%$ more training compute (Section~\ref{sec:h3}; Appendix~\ref{app:h3}), while Wan uses student-curve queries on every hop.

\paragraph{Teacher-query budget.}
The number $m$ of teacher velocity queries per hop is independent of $K$, the number of coefficient vectors used to parameterize curvature.
At generic query positions, every coefficient participates in the same residual, so each query jointly supervises the full velocity profile (Appendix~\ref{app:basis}).
Our default draws one point uniformly from each half of the interval ($m=2$), with equal weights; this is an unbiased estimator of the uniformly integrated squared residual.
Additional queries reuse the same student outputs and objective, increasing teacher cost without changing inference.
On stored H3 teacher paths, the two-query estimation error is only about one eighth of the student's prediction-error energy (Section~\ref{sec:ablations}).

\subsection{Training and inference}
\label{sec:training}
\paragraph{Training on student rollouts.}
Starting from sampled noise, we train successive hops on states produced by the student.
At each hop, one forward pass predicts the curve, local teacher queries define the loss in \eqref{eq:ls}, and the predicted endpoint is detached and carried to the next hop.
After the final hop, the rollout restarts from noise.
This autoregressive procedure trains the student on its own preceding predictions, using prompts and noise without training videos or precomputed trajectories.
Both implementations route gradients within an exact decomposition of \eqref{eq:ls} (Appendix~\ref{app:h3_routing}).
Appendix~\ref{app:hparams} gives the update schedule.

\paragraph{Inference.}
The endpoint identity in \eqref{eq:endpoint} gives the sampling rule
\begin{equation}
  \vx_{k+1}=\vx_k+\Delta_k\vh_\theta(\vx_k,t_k\mid y),
  \qquad k=0,\ldots,N-1.
  \label{eq:sample}
\end{equation}
The curvature head is discarded: its role is to shape the supervision used to learn $\vh_\theta$.
Sampling takes $N$ student evaluations, with no teacher queries or interior-curve evaluation.
Each deployed hop retains the original backbone topology and output dimensions (Appendix~\ref{app:deployment}).

\section{Experiments}
\label{sec:exp}
We test whether \method{} can improve four-step video quality through trajectory supervision alone.
We compare \method{} with LightX2V Turbo on MiniMax-H3 (Section~\ref{sec:h3}) and with PDD on Wan2.1-14B (Section~\ref{sec:wan}).
Ablations then examine how supervision and curve parameterization affect learning within the student's limited capacity (Section~\ref{sec:ablations}).

\subsection{Baselines and evaluation protocol}
\label{sec:protocol}
\paragraph{Baselines.}
PDD~\citep{anonymous2026pdd} releases neither a training recipe nor weights for MiniMax-H3, so on H3 we compare with LightX2V Turbo~\citep{modeltc2026h3turbo}, a state-of-the-art open-source four-step DMD accelerator, using its released v1.1 LoRA and sampler settings.
This cross-paradigm comparison tests whether trajectory supervision can retain a DMD student's static quality while producing more varied, natural-looking videos.
On Wan2.1-14B, PDD is the best-performing trajectory-only method, so we reproduce it under the same training setting as \method{} for a controlled within-paradigm comparison.

\paragraph{Evaluation protocol.}
Following prior work that treats human judgment as the primary signal~\citep{polyak2024moviegen,lin2025apt}, our primary evidence is blinded human votes and GPT-6 Astra judgments, since advanced multimodal LLM judges agree with human preference as well as or better than specialized video metrics~\citep{liu2025uve}.
On Wan2.1-14B, we report VBench~\citep{huang2024vbench} as a reference: its scores saturate on strong models~\citep{zheng2025vbench2,he2024videoscore,inbasekar2026worldjen}, and its rankings can diverge from human preference~\citep{fan2025phaseddmd,han2025videobench}.
Astra uses 100 prompts for both comparisons, fixed before video generation: 50 randomly sampled held-out ViMix live-action captions and 50 Claude-generated action prompts; Wan receives only their video descriptions, and each model generates three videos per prompt with shared seeds.
Astra, from sampled frames, and human raters, from full playback, compare a prompt's videos as two rows of three, one row per model, with model identities hidden and row placement balanced.
Astra scores each model on four 1--4 scales:
\emph{dynamic quality} (motion realism, not motion magnitude), \emph{static quality} (per-frame detail and structure, excluding temporal consistency), \emph{naturalness} (color, materials, lighting, and visual rendition), and \emph{diversity} (variation across seeds that respects the prompt).
Raters choose the preferred row or a tie, considering visual quality, motion, naturalness, and diversity.
Appendices~\ref{app:eval} and~\ref{app:gpt_astra_prompt} give the full protocol and rubric.

\subsection{MiniMax-H3: comparison with LightX2V Turbo}
\label{sec:h3}
\paragraph{Setup.}
We distill MiniMax-H3~\citep{minimaxh3}, a 33B joint audio-video model, using prompts and noise, without training videos or a distribution-matching objective.
Our training prompts mix ViMix~\citep{yang2025vimix}, OpenVid~\citep{nan2025openvid}, and VidProM~\citep{wang2024vidprom}, rewritten into H3's official prompt format using its released rewriter.
We train rank-128 LoRA adapters per hop for 150 updates at batch size 512.
The first three hops use $K{=}8$ Legendre terms and two stratified teacher queries on the student curve; the final hop instead regresses to an eight-step teacher-marched endpoint (Appendix~\ref{app:h3}).
The final hop is far smoother than the noise-end hop (Section~\ref{sec:target}), and we empirically found that marching only this hop improves visual detail over two queries on every hop at about 20\% more training compute.
Inference discards the curvature outputs and uses four network evaluations.
Appendix~\ref{app:hparams} specifies the model-specific recipes.
Both students generate 124-frame videos at $768{\times}1344$ resolution.

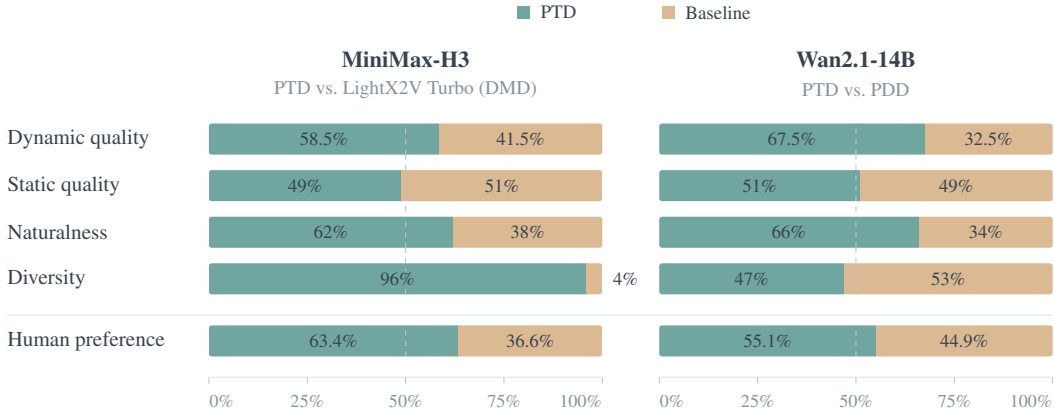
\begin{figure}[!ht]
  \centering
  \resizebox{\linewidth}{!}{\begingroup%
\definecolor{ptdInk}{HTML}{38434E}%
\definecolor{ptdMuted}{HTML}{87939F}%
\definecolor{ptdTeal}{HTML}{70A69F}%
\definecolor{ptdOrange}{HTML}{DBB993}%
\definecolor{ptdRule}{HTML}{DDE3E8}%
\newcommand{\ptdShareLabel}[1]{%
  \pgfmathtruncatemacro{\shareWhole}{floor(#1/10)}%
  \pgfmathtruncatemacro{\shareDecimal}{mod(#1,10)}%
  \shareWhole\ifnum\shareDecimal>0.\shareDecimal\fi\%
}%
\newcommand{\ptdOutcomeRow}[4]{%
  \pgfmathsetmacro{\ptdShare}{100*(#2+.5*#3)/(#2+#3+#4)}%
  \pgfmathsetmacro{\baselineShare}{100-\ptdShare}%
  \pgfmathtruncatemacro{\leftTenths}{round(10*\ptdShare)}%
  \pgfmathtruncatemacro{\rightTenths}{1000-\leftTenths}%
  \begin{scope}
    \clip[rounded corners=1.1pt] (0,#1-6) rectangle (152,#1+6);
    \fill[ptdTeal] (0,#1-6) rectangle (1.52*\ptdShare,#1+6);
    \fill[ptdOrange] (1.52*\ptdShare,#1-6) rectangle (152,#1+6);
    \draw[ptdMuted!55,dash pattern=on 2pt off 2pt,line width=.45pt]
      (76,#1-6) -- (76,#1+6);
  \end{scope}
  \node[label,fill=ptdTeal,inner xsep=.65pt,inner ysep=.3pt,
    font=\rmfamily\upshape\fontsize{7.2}{8}\selectfont]
    at (.76*\ptdShare,#1) {\ptdShareLabel{\leftTenths}};
  \ifnum\rightTenths<80
    \node[label,anchor=west,font=\rmfamily\upshape\fontsize{7.2}{8}\selectfont]
      at (156,#1) {\ptdShareLabel{\rightTenths}};
  \else
    \node[label,fill=ptdOrange,inner xsep=.65pt,inner ysep=.3pt,
      font=\rmfamily\upshape\fontsize{7.2}{8}\selectfont]
      at (152-.76*\baselineShare,#1) {\ptdShareLabel{\rightTenths}};
  \fi
}%
\newcommand{\ptdPanelAxis}{%
  \draw[ptdRule,line width=.4pt] (0,2) -- (152,2);
  \foreach \xx/\score in {0/0,38/25,76/50,114/75,152/100} {
    \draw[ptdMuted!55,line width=.35pt] (\xx,2) -- (\xx,-.5);
    \ifnum\score=0\def\ptdTickAnchor{west}\else
      \ifnum\score=100\def\ptdTickAnchor{east}\else\def\ptdTickAnchor{center}\fi
    \fi
    \node[label,anchor=\ptdTickAnchor,text=ptdMuted,font=\rmfamily\upshape\fontsize{7}{8}\selectfont]
      at (\xx,-7) {\score\%};
  }
}%
\begin{tikzpicture}[x=1pt,y=1pt,text=ptdInk,
  font=\rmfamily\upshape\fontsize{8}{9}\selectfont,
  label/.style={inner sep=0pt}]
  \path[use as bounding box] (0,-39) rectangle (408,122);
  \foreach \xx/\cc/\name in {197/ptdTeal/PTD,253/ptdOrange/Baseline} {
    \fill[\cc] (\xx,114.5) rectangle (\xx+5,119.5);
    \node[label,anchor=west,font=\rmfamily\upshape\fontsize{7.3}{8.5}\selectfont]
      at (\xx+9,117) {\name};
  }
  \foreach \yy/\metric in {68/Dynamic quality,50/Static quality,32/Naturalness,14/Diversity,-10/Human preference} {
    \node[label,anchor=west,font=\rmfamily\upshape\fontsize{8.2}{9.2}\selectfont]
      at (0,\yy) {\metric};
  }
  \draw[ptdRule,line width=.4pt] (0,0) -- (404,0);
  \begin{scope}[xshift=78pt]
    \node[label,font=\rmfamily\upshape\bfseries\fontsize{8.8}{10}\selectfont]
      at (76,99) {MiniMax-H3};
    \node[label,text=ptdMuted,font=\rmfamily\upshape\fontsize{7.5}{8.5}\selectfont]
      at (76,87) {PTD vs.\ LightX2V Turbo (DMD)};
    \draw[ptdMuted!50,dash pattern=on 2pt off 2pt,line width=.45pt]
      (76,8) -- (76,74);
    \ptdOutcomeRow{68}{45}{27}{28}
    \ptdOutcomeRow{50}{6}{86}{8}
    \ptdOutcomeRow{32}{28}{68}{4}
    \ptdOutcomeRow{14}{93}{6}{1}
    \ptdOutcomeRow{-10}{45}{14}{23}
    \begin{scope}[yshift=-26pt]
      \ptdPanelAxis
    \end{scope}
  \end{scope}
  \begin{scope}[xshift=252pt]
    \node[label,font=\rmfamily\upshape\bfseries\fontsize{8.8}{10}\selectfont]
      at (76,99) {Wan2.1-14B};
    \node[label,text=ptdMuted,font=\rmfamily\upshape\fontsize{7.5}{8.5}\selectfont]
      at (76,87) {PTD vs.\ PDD};
    \draw[ptdMuted!50,dash pattern=on 2pt off 2pt,line width=.45pt]
      (76,8) -- (76,74);
    \ptdOutcomeRow{68}{47}{41}{12}
    \ptdOutcomeRow{50}{12}{78}{10}
    \ptdOutcomeRow{32}{44}{44}{12}
    \ptdOutcomeRow{14}{36}{22}{42}
    \ptdOutcomeRow{-10}{33}{21}{25}
    \begin{scope}[yshift=-26pt]
      \ptdPanelAxis
    \end{scope}
  \end{scope}
\end{tikzpicture}%
\endgroup%
}
  \caption{\textbf{\method{} improves naturalness over both baselines and receives majority human preference shares.}
  The first four rows show Astra-derived outcomes over 100 prompts, with ties split equally; dashed lines mark 50\%.
  Outcomes use scores averaged over four calls per prompt, with differences below $0.25$ in magnitude counted as ties.
  Human preference is a separate holistic vote (Appendix~\ref{app:human}).
  Both evaluations use H3 PTD at 150 updates and Wan PTD at 64 versus PDD at 96.
  Appendix Table~\ref{tab:h3_evaluation} reports the automated scores.}
  \label{fig:astra_main}
  \label{fig:h3_main}
\end{figure}

\paragraph{Results.}
\method{} significantly improves Astra diversity from $1.55$ to $2.81$ and naturalness from $2.98$ to $3.09$ while matching LightX2V Turbo in dynamic and static quality (Figure~\ref{fig:astra_main}; Appendix Table~\ref{tab:h3_evaluation}).
Both gains hold on the live-action and action prompt groups (Appendix~\ref{app:astra_breakdown}).
Blinded raters give \method{} a 63.4\% preference share against LightX2V Turbo.
Figure~\ref{fig:teaser} illustrates natural carving motion and variation across seeds in snowboarding.

\subsection{Wan2.1-14B: comparison with PDD}
\label{sec:wan}
\paragraph{Setup.}
We fully fine-tune Wan2.1-T2V-14B~\citep{wan2025} on randomly sampled ViMix prompts, using two stratified student-curve queries on all four hops.
We reproduce PDD strictly following its reported training hyperparameters, with the same backbone, frozen teacher and guidance, ViMix training prompts, global batch size 256, and constant learning rate $10^{-5}$ as \method{}.
Both methods generate 81-frame videos at $480{\times}832$ resolution.

\paragraph{Results.}
Under this shared setting, \method{} outperforms PDD: it significantly improves Astra dynamic quality ($2.43$ vs.\ $2.16$) and naturalness ($2.82$ vs.\ $2.60$) while matching static quality (Figure~\ref{fig:astra_main}; Appendix Table~\ref{tab:h3_evaluation}), and it receives a 55.1\% human preference share.
After only 64 updates, \method{} also exceeds the overall VBench score of all three evaluated checkpoints of our PDD reproduction, at 64, 96, and 256 updates (Table~\ref{tab:main}; Appendix Table~\ref{tab:wan_checkpoint_selection}).
We attribute the gap between our PDD reproduction and PDD's published VBench scores to differences in training data and evaluation prompts, together with evaluation noise.
\begin{table}[!ht]
  \centering
  \caption{\textbf{VBench and diversity on Wan2.1-14B.} Under the same training setting, \method{} at 64 updates outperforms the best checkpoint of our PDD reproduction (96 updates) in overall, quality, and semantic VBench scores; bold compares these two rows. Gray scores are published reference results from \citet{anonymous2026pdd}.}
  \label{tab:main}
  \small
  \newcommand{\pddreported}[1]{\textcolor{black!55}{#1}}
  \setlength{\tabcolsep}{3pt}
  \renewcommand{\arraystretch}{1.10}
  \begin{tabular*}{\linewidth}{@{\extracolsep{\fill}}lcrrrrr@{}}
    \toprule
    \multirow{2}{*}{Method} & \multirow{2}{*}{NFE} & \multicolumn{3}{c}{VBench $\uparrow$} & \multicolumn{2}{c}{Diversity $\uparrow$} \\
    \cmidrule(lr){3-5}\cmidrule(l){6-7}
    & & Overall & Quality & Semantic & V-JEPA 2 & VideoMAE \\
    \midrule
    \multicolumn{7}{@{}l}{\textit{Published reference results}} \\
    UniPC$^{*}$ (Teacher) & $50{\times}2$ & \pddreported{83.90} & \pddreported{84.56} & \pddreported{81.24} & \pddreported{0.1263} & \pddreported{0.0250} \\
    \midrule
    rCM$^{\dagger}$ & 4 & \pddreported{84.92} & \pddreported{85.43} & \pddreported{82.88} & \pddreported{---} & \pddreported{---} \\
    AnyFlow$^{*}$ & 4 & \pddreported{84.95} & \pddreported{85.70} & \pddreported{81.92} & \pddreported{0.0786} & \pddreported{0.0130} \\
    DMD2$^{**}$ (FastGen) & 4 & \pddreported{84.40} & \pddreported{85.16} & \pddreported{81.34} & \pddreported{0.0568} & \pddreported{0.0095} \\
    PDD$^{\mathrm{short}}$ (Midpoint) & 4 & \pddreported{84.92} & \pddreported{85.71} & \pddreported{81.77} & \pddreported{0.0791} & \pddreported{0.0125} \\
    PDD$^{\mathrm{long}}$ (Midpoint) & 4 & \pddreported{84.69} & \pddreported{85.69} & \pddreported{80.71} & \pddreported{0.0846} & \pddreported{0.0126} \\
    \midrule
    \multicolumn{7}{@{}l}{\textit{Our best evaluated checkpoints}} \\
    PDD @96 (reproduction) & 4 & 83.67 & 85.24 & 77.40 & \textbf{0.1340} & \textbf{0.0254} \\
    \textbf{\method{} @64 (ours)} & 4 & \textbf{84.57} & \textbf{85.44} & \textbf{81.10} & 0.1220 & 0.0180 \\
    \bottomrule
  \end{tabular*}
  \par\vspace{3pt}
  \begin{minipage}{\linewidth}
    \scriptsize\raggedright
    $^{*}$Official checkpoints re-evaluated by the PDD authors; $^{**}$their FastGen implementation.
    $^{\dagger}$rCM uses proprietary prompts; other published rows use Self Forcing prompts.
    PDD short/long denotes 200/3{,}000 training iterations. NFE (number of function evaluations) includes guidance passes.
    Published rows use different training budgets and protocols. Diversity is cosine distance in V-JEPA 2~\citep{assran2025vjepa2} and VideoMAE V2~\citep{wang2023videomaev2} feature space; the teacher uses the UniPC solver~\citep{zhao2023unipc}; rCM diversity is not reported in the cited comparison.
  \end{minipage}
\end{table}

\begin{figure}[!ht]
  \centering
  \resizebox{\linewidth}{!}{\begingroup%
\definecolor{curveInk}{HTML}{38434E}%
\definecolor{curveMuted}{HTML}{87939F}%
\definecolor{curveTeal}{HTML}{70A69F}%
\definecolor{curveOrange}{HTML}{DBB993}%
\definecolor{curveRule}{HTML}{E7EAEC}%
\begin{tikzpicture}[x=1pt,y=1pt,text=curveInk,
  font=\rmfamily\upshape\fontsize{8}{9}\selectfont,
  label/.style={inner sep=0pt},
  series/.style={line width=1.15pt,line cap=round,line join=round}]
  \path[use as bounding box] (0,0) rectangle (396,104);
  \node[label,anchor=west,font=\rmfamily\upshape\fontsize{8.3}{9.3}\selectfont]
    at (0,99) {Teacher-aligned bend projection (\%)};
  \draw[series,curveTeal] (295,99) -- (307,99);
  \fill[curveTeal] (301,99) circle (1.7pt);
  \node[label,anchor=west] at (312,99) {PTD};
  \draw[series,curveOrange,dash pattern=on 3pt off 1.5pt] (350,99) -- (362,99);
  \draw[curveOrange,line width=.85pt] (354.3,97.3) rectangle (357.7,100.7);
  \node[label,anchor=west] at (367,99) {PDD};

  \foreach \offset/\title in {0/High noise (hop 0),204/Final (hop 3)} {
    \begin{scope}[xshift=\offset pt]
      \node[label,font=\rmfamily\bfseries\upshape\fontsize{8.5}{9.5}\selectfont]
        at (107,83) {\title};
      \foreach \v in {0,2,4,6} {
        \draw[curveRule,line width=.35pt] (26,20+7.5*\v) -- (188,20+7.5*\v);
        \node[label,anchor=east,text=curveMuted,font=\rmfamily\upshape\fontsize{7.2}{8}\selectfont]
          at (21,20+7.5*\v) {\v};
      }
      \draw[curveMuted!70,line width=.4pt] (26,70) -- (26,20) -- (188,20);
      \draw[curveMuted!65,dash pattern=on 2pt off 2pt,line width=.5pt]
        (53,20) -- (53,70);
      \foreach \step/\xx in {32/26,64/53,128/107,192/161,224/188} {
        \draw[curveMuted!70,line width=.4pt] (\xx,20) -- (\xx,17);
        \node[label,text=curveMuted,font=\rmfamily\upshape\fontsize{7.2}{8}\selectfont]
          at (\xx,10) {\step};
      }
    \end{scope}
  }
  \node[label] at (198,0) {Training updates};
  \draw[series,curveTeal] plot coordinates {
    (26,27.5) (53,30.5) (80,35.75) (107,44.75) (134,50) (161,54.5) (188,60.5)};
  \draw[series,curveOrange,dash pattern=on 3pt off 1.5pt] plot coordinates {
    (26,21.5) (53,22.25) (80,24.5) (107,26.75) (134,26) (161,28.25) (188,30.5)};
  \foreach \xx/\yy in {26/27.5,53/30.5,80/35.75,107/44.75,134/50,161/54.5,188/60.5} {
    \fill[curveTeal] (\xx,\yy) circle (1.7pt);
  }
  \foreach \xx/\yy in {26/21.5,53/22.25,80/24.5,107/26.75,134/26,161/28.25,188/30.5} {
    \draw[curveOrange,line width=.85pt] (\xx-1.7,\yy-1.7) rectangle (\xx+1.7,\yy+1.7);
  }
  \draw[series,curveTeal] plot coordinates {
    (230,28.25) (257,36.5) (284,43.25) (311,48.5) (338,53.75) (365,59.75) (392,65.75)};
  \draw[series,curveOrange,dash pattern=on 3pt off 1.5pt] plot coordinates {
    (230,29.75) (257,40.25) (284,46.25) (311,51.5) (338,55.25) (365,59.75) (392,63.5)};
  \foreach \xx/\yy in {230/28.25,257/36.5,284/43.25,311/48.5,338/53.75,365/59.75,392/65.75} {
    \fill[curveTeal] (\xx,\yy) circle (1.7pt);
  }
  \foreach \xx/\yy in {230/29.75,257/40.25,284/46.25,311/51.5,338/55.25,365/59.75,392/63.5} {
    \draw[curveOrange,line width=.85pt] (\xx-1.7,\yy-1.7) rectangle (\xx+1.7,\yy+1.7);
  }
\end{tikzpicture}%
\endgroup%
}
  \caption{\textbf{Learning teacher-aligned curvature on Wan2.1-14B.}
  Bend projection normalized by teacher bend energy (eight prompts).
  PTD learns a larger aligned component on the high-noise first hop; final-hop projections remain close.
  Vertical dashed lines mark 64 updates, the evaluated \method{} checkpoint.}
  \label{fig:wan_curve_learning}
\end{figure}
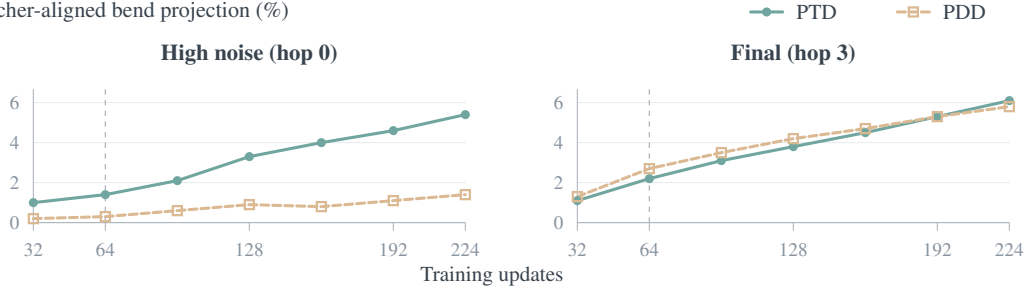

\paragraph{Learning within-hop trajectory shape.}
On eight VBench prompts with a shared 128-slot teacher reference, we measure how well each student learns the teacher's within-hop bend (Figure~\ref{fig:wan_curve_learning}).
On the high-noise first hop, \method{} has a larger bend magnitude ratio $r$ and a higher direction cosine than PDD at every shared update count from 32 to 224; at 64 updates, $r$ is $0.081$ versus $0.024$, and the teacher-aligned projection is $0.014$ versus $0.003$.
Final-hop projections remain close.
\method{}'s gain is largest on this high-noise hop, where trajectory curvature is high and Section~\ref{sec:target} motivates querying the student curve rather than regressing to teacher-integrated targets.
Appendix~\ref{sec:trajectory_learning} gives all-hop results and metric definitions.

\subsection{Ablations}
\label{sec:ablations}
\begin{figure}[!ht]
  \centering
  \includegraphics[width=\linewidth]{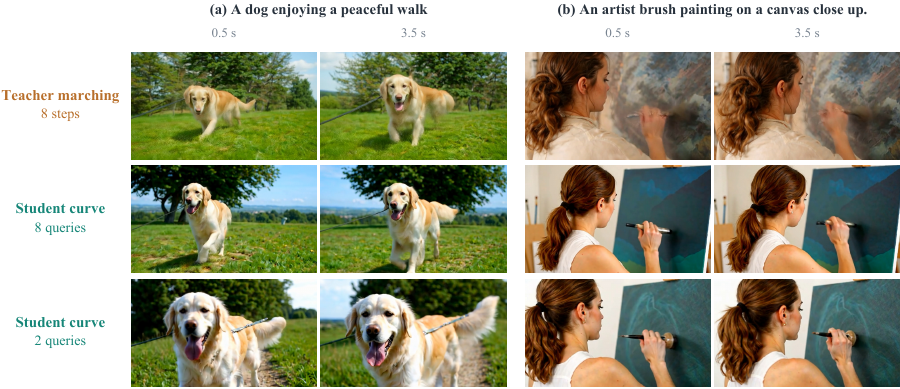}
  \caption{\textbf{Student-curve supervision preserves local detail.}
  Three Wan2.1-1.3B students after 40 training updates, all using four-step inference, shown at the same two timestamps per prompt.
  Teacher marching on every hop softens the dog's coat and the painter's hand and brush; both student-curve variants retain distinct detail.
  Step and query counts refer to teacher supervision per training hop.}
  \label{fig:marching_qualitative}
\end{figure}

\paragraph{Student-curve queries versus teacher marching.}
We compare three supervision schemes on Wan2.1-1.3B: an eight-step teacher-marched endpoint target on every hop, eight teacher queries on the student curve, and our two-query default.
All three students are trained for 40 updates and use four evaluations at inference.
The marching baseline integrates the teacher from each carried hop input and trains $\vh_\theta$ to predict the resulting mean velocity; the student-curve variants jointly train $\vh_\theta$ and the curve coefficients with the loss in \eqref{eq:ls}.
Marching on every hop visibly softens local detail around the dog's coat and legs and the painter's hand and brush, whereas both student-curve variants retain distinct structure (Figure~\ref{fig:marching_qualitative}).
As \Figref{fig:student_path_supervision} anticipates, a more accurate teacher target is harder for a few-step student to learn, especially on high-noise hops, where teacher trajectories bend most (Section~\ref{sec:target}).
Student-curve queries give learnable guidance on these curved hops; on the smoother final hop, we empirically found that marching improves H3's visual detail (Section~\ref{sec:h3}).

\begin{figure}[!ht]
  \centering
  \begin{minipage}[c]{0.5\linewidth}
    \centering
    \resizebox{\linewidth}{!}{\begingroup%
\definecolor{readInk}{HTML}{27313C}%
\definecolor{readMuted}{HTML}{788390}%
\definecolor{readTeal}{HTML}{168678}%
\definecolor{readAmber}{HTML}{C98A4A}%
\definecolor{readRule}{HTML}{DEE4E8}%
\begin{tikzpicture}[x=1pt,y=1pt,text=readInk,
  font=\rmfamily\fontsize{7.5}{9}\selectfont,
  label/.style={inner sep=0pt}]
  \path[use as bounding box] (0,0) rectangle (198,110);
  \fill[readMuted!32] (28,98) rectangle (35,104);
  \node[label,anchor=base west] at (39,98.5) {Student};
  \fill[readTeal] (78,98) rectangle (85,104);
  \node[label,anchor=base west] at (89,98.5) {Two queries};
  \fill[readAmber] (138,98) rectangle (145,104);
  \node[label,anchor=base west] at (149,98.5) {Three queries};
  \node[label,rotate=90] at (6,52) {Relative MSE (\%)};
  \foreach \v in {0,50,100} {
    \draw[readRule,line width=.35pt] (34,24+.56*\v) -- (196,24+.56*\v);
    \node[label,anchor=east,text=readMuted] at (29,24+.56*\v) {\v};
  }
  \draw[readMuted!70,line width=.45pt] (34,83) -- (34,24) -- (196,24);
  \foreach \x/\hop/\ratio/\txt/\three/\threeTxt in {62/0/12.606097/12.6/6.646742/6.6,117/1/13.596491/13.6/6.052632/6.1,172/2/13.333333/13.3/5.833333/5.8} {
    \fill[readMuted!32] (\x-23,24) rectangle (\x-13,80);
    \fill[readTeal] (\x-5,24) rectangle (\x+5,24+.56*\ratio);
    \fill[readAmber] (\x+13,24) rectangle (\x+23,24+.56*\three);
    \node[label,anchor=south,text=readTeal,
      font=\rmfamily\bfseries\fontsize{7}{8.4}\selectfont]
      at (\x,28+.56*\ratio) {\txt\%};
    \node[label,anchor=south,text=readAmber,
      font=\rmfamily\bfseries\fontsize{7}{8.4}\selectfont]
      at (\x+18,28+.56*\three) {\threeTxt\%};
    \node[label] at (\x,13) {Hop \hop};
  }
\end{tikzpicture}%
\endgroup%
}
  \end{minipage}\hfill
  \begin{minipage}[c]{0.46\linewidth}
    \captionsetup{font=small,justification=raggedright,singlelinecheck=false}
    \caption{\textbf{Two queries give a small target-estimation error.}
    On video hops 0--2, two-query quadrature MSE is about one eighth of student MSE; three queries reduce it further.
    Each hop's student MSE sets the reference scale (100\%).
    Quadrature and student errors use separate banks of 100 and 96 prompts, respectively.
    Appendix~\ref{app:read_ablation} gives the protocol.}
    \label{fig:two_read_target}
  \end{minipage}
\end{figure}
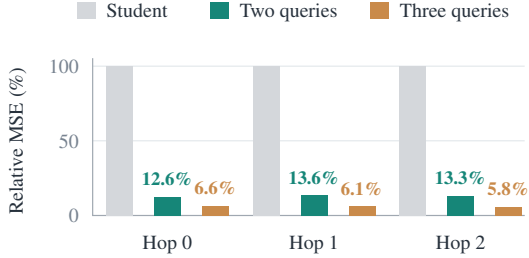

\paragraph{Accuracy of two-query target estimation.}
Our default uses two stratified queries per student-curve-supervised hop.
On stored H3 teacher trajectories, we compare the average of one velocity query per half-hop with the endpoint-derived mean.
For hops 0--2, the two-query quadrature MSE is only $12.6\%$, $13.6\%$, and $13.3\%$ of the student's prediction-error energy (Figure~\ref{fig:two_read_target}), so two queries already give an accurate target.
A third query lowers these ratios to $6.6\%$, $6.1\%$, and $5.8\%$ at $50\%$ more teacher queries per hop (Appendix~\ref{app:read_ablation}).

\begin{figure}[!ht]
  \centering
  \begin{minipage}[c]{0.5\linewidth}
    \centering
    \resizebox{\linewidth}{!}{\begingroup%
\definecolor{probeInk}{HTML}{27313C}%
\definecolor{probeMuted}{HTML}{788390}%
\definecolor{probeBlue}{HTML}{3569B8}%
\definecolor{probeTeal}{HTML}{168678}%
\definecolor{probeOrange}{HTML}{CC7B2D}%
\definecolor{probeRule}{HTML}{DEE4E8}%
\begin{tikzpicture}[x=1pt,y=1pt,text=probeInk,
  font=\rmfamily\fontsize{7.5}{9}\selectfont,
  label/.style={inner sep=0pt},
  series/.style={line width=.8pt,line cap=round,line join=round}]
  \path[use as bounding box] (0,0) rectangle (198,110);

  \draw[series,probeBlue] (28,104) -- (39,104);
  \fill[probeBlue] (33.5,104) circle (1.5pt);
  \node[label,anchor=west] at (43,104) {Fit: noise-end};
  \draw[series,probeTeal] (120,104) -- (131,104);
  \fill[probeTeal] (124,102.5) rectangle (127,105.5);
  \node[label,anchor=west] at (135,104) {Fit: final};
  \draw[series,probeMuted,dash pattern=on 3pt off 2pt] (28,93) -- (39,93);
  \fill[probeMuted] (32,91.5) rectangle (35,94.5);
  \node[label,anchor=west] at (43,93) {Probe: ridge};
  \draw[series,probeOrange,dash pattern=on 3pt off 2pt] (120,93) -- (131,93);
  \fill[probeOrange] (125.5,93) circle (1.5pt);
  \node[label,anchor=west] at (135,93) {Probe: Adam};

  \node[label,rotate=90] at (5,51) {Explained variation (\%)};
  \fill[probeBlue!5] (151,22) rectangle (159,81);
  \foreach \v in {0,25,50,75,100} {
    \draw[probeRule,line width=.35pt] (38,22+.56*\v) -- (194,22+.56*\v);
    \node[label,anchor=east,text=probeMuted,font=\rmfamily\fontsize{7}{8}\selectfont]
      at (33,22+.56*\v) {\v};
  }
  \draw[probeMuted!70,line width=.45pt] (38,81) -- (38,22) -- (194,22);
  \draw[probeMuted!70,dash pattern=on 2pt off 2pt,line width=.5pt]
    (155,22) -- (155,82);
  \foreach \k in {1,2,3,4,8,16} {
    \pgfmathsetmacro{\xx}{38+39*ln(\k)/ln(2)}
    \draw[probeMuted!70,line width=.45pt] (\xx,22) -- (\xx,19);
    \node[label] at (\xx,12) {\k};
  }
  \node[label] at (116,1.5) {Curve terms $K$};

  \draw[series,probeBlue] plot coordinates {
    (38,36) (77,47.76) (99.81354,56.16) (116,61.76) (155,72.4) (194,76.88)};
  \foreach \k/\v in {1/25,2/46,3/61,4/71,8/90,16/98} {
    \pgfmathsetmacro{\xx}{38+39*ln(\k)/ln(2)}
    \fill[probeBlue] (\xx,22+.56*\v) circle (1.25pt);
  }
  \draw[series,probeTeal] plot coordinates {
    (38,64.56) (77,72.4) (99.81354,75.2) (116,76.32) (155,77.44) (194,78)};
  \foreach \k/\v in {1/76,2/90,3/95,4/97,8/99,16/100} {
    \pgfmathsetmacro{\xx}{38+39*ln(\k)/ln(2)}
    \fill[probeTeal] (\xx-1.2,22+.56*\v-1.2) rectangle (\xx+1.2,22+.56*\v+1.2);
  }
  \node[label,anchor=west,text=probeBlue,fill=white,inner sep=.5pt,
    font=\rmfamily\bfseries\fontsize{7.5}{9}\selectfont] at (170,68) {90\%};
  \node[label,anchor=south east,text=probeTeal,
    font=\rmfamily\bfseries\fontsize{7.5}{9}\selectfont] at (151,81) {99\%};

  \draw[series,probeMuted,dash pattern=on 3pt off 2pt] plot coordinates {
    (38,46.08) (77,47.2) (99.81354,47.2) (116,47.2) (155,47.2)};
  \foreach \k/\v in {1/43,2/45,3/45,4/45,8/45} {
    \pgfmathsetmacro{\xx}{38+39*ln(\k)/ln(2)}
    \fill[probeMuted] (\xx-1.2,22+.56*\v-1.2) rectangle (\xx+1.2,22+.56*\v+1.2);
  }
  \draw[series,probeOrange,dash pattern=on 3pt off 2pt] plot coordinates {
    (38,48.88) (77,49.44) (99.81354,50) (116,50) (155,50)};
  \foreach \k/\v in {1/48,2/49,3/50,4/50} {
    \pgfmathsetmacro{\xx}{38+39*ln(\k)/ln(2)}
    \fill[probeOrange] (\xx,22+.56*\v) circle (1.25pt);
  }
  \draw[probeOrange,line width=.95pt] (155,49.44) -- (155,50);
  \draw[probeOrange,line width=.95pt] (152.5,49.44) -- (157.5,49.44);
  \draw[probeOrange,line width=.95pt] (152.5,50) -- (157.5,50);
  \node[label,text=probeOrange,fill=white,inner sep=.5pt,
    font=\rmfamily\bfseries\fontsize{7.5}{9}\selectfont] at (155,59) {49--50\%};
  \node[label,text=probeMuted,fill=white,inner sep=.5pt,
    font=\rmfamily\bfseries\fontsize{7.5}{9}\selectfont] at (155,38) {45\%};
\end{tikzpicture}%
\endgroup%
}
  \end{minipage}\hfill
  \begin{minipage}[c]{0.46\linewidth}
    \captionsetup{font=small,justification=raggedright,singlelinecheck=false}
    \caption{\textbf{Curve capacity far exceeds what student features can predict.}
    Offline Legendre fits (8 teacher trajectories) and final-hop linear probes (500 held-out prompts) use separate H3 banks.
    Solid and dashed lines denote fits and predictions, respectively.
    The band marks the default $K=8$, where the Adam range spans parameterizations; probes cover $K\le8$.
    See Appendix~\ref{app:curve_probes}.}
    \label{fig:curve_probe}
  \end{minipage}
\end{figure}
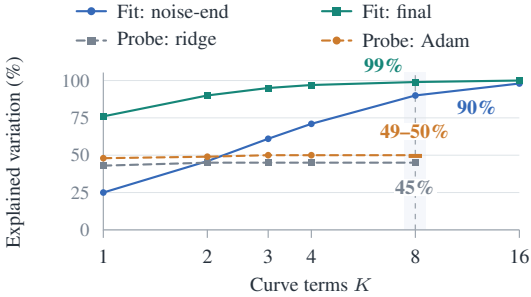

\begin{samepage}
\paragraph{Curve capacity and learnable curvature.}
Figure~\ref{fig:curve_probe} separates offline H3 trajectory fits from linear probes on frozen student features.
Raising $K$ from 2 to 8 lifts the offline explained fraction from $46\%$ to $90\%$ on the strongly curved hop nearest noise and from $90\%$ to $99\%$ on the smoother final hop, so the curve family can represent teacher trajectories almost exactly.
Student features predict far less: held-out probes explain $45$--$50\%$ of final-hop variation for every $K\in\{2,3,4,8\}$ and only $7$--$11\%$ on the first hop.
The learned curve is correspondingly a shrunken bend aligned with the teacher's (Section~\ref{sec:wan}), which keeps the queried targets learnable.
Since extra terms raise this ceiling without reducing predictability, we adopt $K{=}8$ as the default; Appendix~\ref{app:curve_probes} details basis conditioning, optimization, and protocols.\par
\end{samepage}

\section{Conclusion}
\label{sec:conclusion}
We presented Parametric Trajectory Distillation, approaching few-step video distillation as a capacity allocation problem.
Teacher queries along student-predicted polynomial paths jointly supervise a mean-velocity output and a curvature head.
This training scaffold is removed at inference, retaining the original backbone architecture and four-step sampling.
Using trajectory supervision alone, \method{} distills the 33B MiniMax-H3 model into a four-step LoRA student that significantly improves Astra diversity and naturalness over the state-of-the-art LightX2V Turbo while matching its dynamic and static quality.
On Wan2.1-14B, \method{} sets a new state of the art for trajectory distillation, outperforming PDD, the best-performing trajectory-only method on this model, under the same training setting, with significant gains in Astra dynamic quality and naturalness.
Blinded human votes give \method{} 63.4\% and 55.1\% preference shares against LightX2V Turbo and PDD.
Our analyses show that the best supervision target varies along the trajectory: teacher marching on every hop softens fine detail, whereas queries along the student's own curve retain it.
On the smoother final hop, we empirically found that a teacher-marched target improves detail.
Although the curve family can fit teacher trajectories almost exactly, the student learns a shrunken, teacher-aligned curve that keeps its targets learnable.
These results point to a simple principle for few-step video distillation: match supervision to what the student can predict.

\newpage
\subsubsection*{AI use statement}
In this work, we used generative AI tools (large language model coding assistants) to write and refactor experiment, probing, and evaluation code, to draft and edit the manuscript text, and to organize the related-work survey.
We used a large language model (Claude) to generate the 50 action prompts in our 100-prompt evaluation set; these prompts were fixed before video generation and were not edited after judging (Appendix~\ref{app:eval}).
We also used GPT-6 Astra as an automated judge for the four-dimensional video evaluation; these judgments are not human ratings.
Experimental measurements were obtained by the authors, including the automated evaluation, or are explicitly attributed to prior work.
All AI-assisted code was reviewed and tested by the authors, and all AI-drafted text was checked against our experimental logs by the authors.
We take responsibility for the final content of this work, including text, claims, and artifacts produced with the aid of generative AI.

\subsubsection*{Reproducibility statement}
Section~\ref{sec:method} and Appendix~\ref{app:hparams} specify the hop grid, the parametric curve, the query positions and velocity least-squares objective, the backbone-specific recipes (per-hop LoRA placement and ranks for MiniMax-H3 and full fine-tuning for Wan2.1-14B), the training setting shared by \method{} and our PDD reproduction, and the optimizer settings.
Appendix~\ref{app:deployment} describes removal of the curvature head and, for MiniMax-H3, adapter merging for inference.
Appendix~\ref{app:eval} describes evaluation prompt construction, the four-dimensional automated evaluation, the Wan2.1-14B VBench protocol, and the blinded human study; Appendix~\ref{app:probes} specifies the trajectory and teacher-query diagnostics, and Appendix~\ref{app:gpt_astra_prompt} gives the full judging rubric.
The trained MiniMax-H3 LoRA adapters and Wan2.1-14B checkpoint, the training and probing code, the evaluation prompts, and the judging scripts will be released.

\bibliography{ptd}
\bibliographystyle{iclr2027_conference}

\clearpage
\appendix
\section{Additional method details}
\label{app:method_details}

\subsection{Endpoint-constrained curves and supervision}
\label{app:basis}
Throughout this appendix we fix a hop $k$ and write $\vh$, $\vc_j$, $\vx_\theta$, and $\vv_\theta$ for $\vh_\theta$, $\vc_{\theta,j}$, $\vx_{\theta,k}$, and $\vv_{\theta,k}$.
With $\sigma=2\tau-1$ the Legendre polynomials are $P_0=1$, $P_1=\sigma$, $P_2=\tfrac12(3\sigma^2-1)$, $P_3=\tfrac12(5\sigma^3-3\sigma)$, and so on; we use $K{=}8$.
The curvature basis functions are $b_j(\tau)=\tau(\tau-1)P_j(2\tau-1)$.
The parameterized trajectory in \eqref{eq:path} is $\vx_\theta(\tau)=\vx_k+\Delta_k[\tau\vh+\sum_{j<K}b_j(\tau)\vc_j]$.
Differentiating with respect to $s=t_k+\tau\Delta_k$ divides by $\Delta_k$ and gives \eqref{eq:tangent} with
\begin{equation}
 d_j(\tau)=\frac{\mathrm d b_j(\tau)}{\mathrm d\tau}
 =(2\tau-1)P_j(\sigma)+2\tau(\tau-1)\frac{\mathrm dP_j(\sigma)}{\mathrm d\sigma}.
\end{equation}
Since $b_j(0)=b_j(1)=0$, the curvature terms leave both endpoints fixed at a given $\vh$ and satisfy $\int_0^1d_j(\tau)\,\mathrm d\tau=0$.
Consequently,
\begin{equation}
 \vx_\theta(1)=\vx_k+\Delta_k\vh,
 \qquad
 \int_0^1\vv_\theta(\tau)\,\mathrm d\tau=\vh.
 \label{eq:curve-endpoint}
\end{equation}
These exact parameterization identities justify retaining $\vh$ as the distilled output and removing the curvature head at inference.

\paragraph{Both heads receive the same residual.}
For fixed, detached teacher queries, define $\vr_i=\vh+\sum_{j<K}d_j(\tau_i)\vc_j-\vu_i$.
Before the implementation's gradient routing (Appendix~\ref{app:h3_routing}), the ordinary output gradients of \eqref{eq:ls} are
\begin{equation}
 \frac{\partial\mathcal L_{\mathrm{PTD}}}{\partial\vh}=2\sum_i\omega_i\vr_i,
 \qquad
 \frac{\partial\mathcal L_{\mathrm{PTD}}}{\partial\vc_j}=2\sum_i\omega_i d_j(\tau_i)\vr_i.
 \label{eq:joint-gradients}
\end{equation}
Thus a local teacher target directly supervises the endpoint output and the curve coefficients through one velocity residual.
The $K$ coefficients are latent-sized vectors emitted by one curvature head.

\paragraph{How the endpoint output learns a mean velocity.}
Fix the query curve $\vx_{\theta_0}$ at the current parameters $\theta_0$ and define the detached teacher field along it as
$\widetilde{\vu}(\tau)=\sg[\teacher(\vx_{\theta_0}(\tau),s(\tau)\mid y)]$.
For uniform queries, or an unbiased stratified estimator of the same integral, \eqref{eq:joint-gradients} gives
\begin{align}
 \mathbb E\!\left[\frac{\partial\mathcal L_{\mathrm{PTD}}}{\partial\vh}\right]
 &=2\int_0^1\left[\vh+\sum_{j<K}d_j(\tau)\vc_j-\widetilde{\vu}(\tau)\right]\,\mathrm d\tau \nonumber\\
 &=2\left[\vh-\int_0^1\widetilde{\vu}(\tau)\,\mathrm d\tau\right].
 \label{eq:mean-output-gradient}
\end{align}
\Eqref{eq:mean-output-gradient} is the output partial derivative of the unnormalized least-squares (LS) objective with query locations held fixed and teacher targets detached, averaged over query positions.
Its target is the teacher velocity averaged along the current student curve; Appendix~\ref{app:family} relates the local velocity residual to the teacher endpoint.
H3's modality normalization subsequently rescales the sampled gradients.

\paragraph{Coverage across query positions.}
For one latent coordinate the least-squares design row is
$\mathbf z(\tau)=[1,d_0(\tau),\ldots,d_{K-1}(\tau)]$.
For $m{=}2$, the design matrix has rank at most two, providing at most two independent constraints on the nine output values for each latent coordinate.
Across random query positions, the basis is fully observable: $d_j$ is a polynomial of degree $j+1$ with a nonzero leading coefficient, so $1,d_0,\ldots,d_{K-1}$ are linearly independent.
Under a query density that is positive throughout $(0,1)$, the expected Gram matrix $\mathbb E[\mathbf z(\tau)^{\mathsf T}\mathbf z(\tau)]$ is therefore positive definite.
For our equal-weight stratified pair, the expectation of \eqref{eq:ls} is exactly the uniform integral of the squared residual, since the two half-intervals together cover $[0,1]$.
This provides supervision of the full profile across training draws.
With $K=8$, the position and velocity polynomials have maximum degrees nine and eight, respectively.

\subsection{Finite-query gradient routing}
\label{app:h3_routing}
Both evaluated PTD recipes use the finite-query LS value in \eqref{eq:ls}, with gradients routed through an exact decomposition on student-curve-supervised hops.
For equal weights, write $\bar d_j=m^{-1}\sum_i d_j(\tau_i)$, $\bar{\vu}=m^{-1}\sum_i\vu_i$, and
\begin{equation}
 \bar{\vr}=\vh+\sum_j\bar d_j\vc_j-\bar{\vu},\qquad
 \widetilde{\vr}_i=\sum_j[d_j(\tau_i)-\bar d_j]\vc_j-(\vu_i-\bar{\vu}).
\end{equation}
Since $\sum_i\widetilde{\vr}_i=0$, the following equality holds for every query set:
\begin{equation}
 \mathcal L_{\mathrm{PTD}}=\|\bar{\vr}\|^2+\frac1m\sum_i\|\widetilde{\vr}_i\|^2.
 \label{eq:ls-decomposition}
\end{equation}
In the first term, the implementation replaces $\sum_j\bar d_j\vc_j$ by its stop-gradient value; we denote the resulting routed objective by $\mathcal L_{\mathrm{route}}$.
Before per-modality loss normalization, this preserves the loss value and the output partial derivative with respect to $\vh$, while changing the curvature gradient:
\begin{equation}
 \frac{\partial\mathcal L_{\mathrm{route}}}{\partial\vh}=2\bar{\vr},\qquad
 \frac{\partial\mathcal L_{\mathrm{route}}}{\partial\vc_j}
 =\frac{2}{m}\sum_i[d_j(\tau_i)-\bar d_j]\widetilde{\vr}_i.
 \label{eq:h3-route-gradients}
\end{equation}
Relative to ordinary LS differentiation, the omitted term is $2\bar d_j\bar{\vr}$.
Because $\bar d_j$ and $\bar{\vr}$ share the sampled query positions, the expected omitted term is $2\operatorname{Cov}(\bar d_j,\bar{\vr})$, since $\mathbb E[\bar d_j]=\int_0^1d_j(\tau)\,\mathrm d\tau=0$ under the stratified queries.
The shared backbone still receives gradients through both $\vh$ and $\vc$.

\subsection{From local velocity fitting to a distilled endpoint}
\label{app:family}
The training target in \method{} is the teacher's local velocity on a parameterized curve.
The relation between this local supervision and the teacher's endpoint follows from the ODE residual, as in the local-dynamics analysis of flow-map distillation~\citep{boffi2024flowmap}.

Fix a hop start $\vx_k$, its conditioning, and the student's outputs.
Let $\varphi_k(\vx_k,s)$ denote the teacher ODE solution from $\vx_k$ at $t_k$, and write $\vx_{\mathrm T}(\tau)=\varphi_k(\vx_k,s(\tau))$.
Define
\begin{equation}
 \vr(\tau)=\vv_\theta(\tau)-\teacher(\vx_\theta(\tau),s(\tau)).
 \label{eq:ode-residual}
\end{equation}
Assume the teacher field is Lipschitz in its state with constant $L_{\mathrm T}$ on a region containing the two paths.
Both paths start at $\vx_k$, and their difference satisfies
\begin{equation}
 \frac{\mathrm d}{\mathrm d\tau}(\vx_\theta-\vx_{\mathrm T})
 =\Delta_k\big[\teacher(\vx_\theta,s)-\teacher(\vx_{\mathrm T},s)+\vr(\tau)\big].
\end{equation}
Gronwall's inequality and Cauchy--Schwarz give
\begin{align}
 \|\vx_k+\Delta_k\vh-\vx_{\mathrm T}(1)\|
 &\le |\Delta_k|\int_0^1
 e^{|\Delta_k|L_{\mathrm T}(1-\tau)}\|\vr(\tau)\|\,\mathrm d\tau \nonumber\\
 &\le |\Delta_k|e^{|\Delta_k|L_{\mathrm T}}
 \left(\int_0^1\|\vr(\tau)\|^2\,\mathrm d\tau\right)^{1/2}.
 \label{eq:endpoint-residual-bound}
\end{align}
Thus, under the Lipschitz condition, the student-curve objective directly controls the distilled endpoint: with equal-weight stratified queries and fixed outputs, the expectation of \eqref{eq:ls} equals $\int_0^1\|\vr(\tau)\|^2\,\mathrm d\tau$, which bounds the error of $\vx_k+\Delta_k\vh$ through \eqref{eq:endpoint-residual-bound} without integrating the teacher.
If the residual vanishes throughout the hop, ODE uniqueness gives $\vx_\theta(\tau)=\vx_{\mathrm T}(\tau)$, and $\vh$ equals the teacher trajectory's average velocity.
The curvature coefficients represent this path during training; the endpoint uses only $\vh$.

\subsection{Relation to Lagrangian flow maps, TVM, PDD, and $\pi$-Flow}
\label{app:relations}
This section places \method{} among flow-map, terminal-velocity, interval-supervision, and policy-based distillation (Table~\ref{tab:relations}).
We keep our convention throughout: generation runs from noise at $t=1$ to data at $t=0$, a map runs from a start time $t$ to a target time $s\le t$, and $\Phi_{t\to s}$ denotes the teacher flow map, so $\Phi_{t_k\to s(\tau)}(\vx_k)=\varphi_k(\vx_k,s(\tau))=\vx_{\mathrm T}(\tau)$ in Appendix~\ref{app:family}.
We rewrite equations from works that order the two times differently in this convention.

\begin{table}[t]
  \centering\footnotesize
  \setlength{\tabcolsep}{3pt}
  \caption{\textbf{\method{} among Lagrangian and related distillation methods.}
  LMD: Lagrangian map distillation~\citep{boffi2024flowmap,boffi2025howto}; TVM: Terminal Velocity Matching~\citep{zhou2026tvm}; $\pi$-Flow~\citep{chen2026piflow}; PDD~\citep{anonymous2026pdd}.
  RK: Runge--Kutta; JVP: Jacobian--vector product.
  \method{}'s entries describe its student-curve loss; H3's final interval uses a teacher march instead (Appendix~\ref{app:h3}).}
  \label{tab:relations}
  \begin{tabular}{@{}>{\raggedright\arraybackslash}p{1.25cm}>{\raggedright\arraybackslash}p{2.9cm}>{\raggedright\arraybackslash}p{3.05cm}>{\raggedright\arraybackslash}p{2.9cm}>{\raggedright\arraybackslash}p{2.75cm}@{}}
    \toprule
    & Velocity target & Target query states & Student velocity & Deployed update \\
    \midrule
    LMD & Frozen teacher & Predicted map states & JVP of a two-time network & Two-time map \\
    TVM & Own EMA velocity, plus flow matching on data & Predicted terminal states & JVP of a two-time network, backpropagated & Two-time map \\
    $\pi$-Flow & Frozen teacher & Detached policy rollout, numerically integrated & Closed-form policy & Policy integrated with substeps \\
    PDD & Teacher midpoint RK target & Teacher solver stages & Discrete interval outputs & Fused interval outputs \\
    \method{} & Frozen teacher & Detached student curve, in closed form & Analytic basis derivative & $\vx_k+\Delta_k\vh_\theta$ \\
    \bottomrule
  \end{tabular}
\end{table}

\paragraph{Lagrangian flow-map distillation.}
The teacher flow map is the unique solution of the Lagrangian equation~\citep{boffi2024flowmap}
\begin{equation}
 \partial_s\Phi_{t\to s}(\vx)=\teacher\big(\Phi_{t\to s}(\vx),s\big),
 \qquad \Phi_{t\to t}(\vx)=\vx.
 \label{eq:lagrangian}
\end{equation}
Lagrangian map distillation (LMD) fits a map $\hat X_{t\to s}$ with $\hat X_{t\to t}(\vx)=\vx$ by penalizing the residual of \eqref{eq:lagrangian} at the map's own predicted states.
In the stop-gradient form of \citet{boffi2025howto}, which avoids the teacher's spatial Jacobian, the objective is
\begin{equation}
 \mathcal L_{\mathrm{LMD}}(\hat X)
 =\mathbb E_{(t,s)\sim w}\,\mathbb E_{\vx\sim\rho_t}
 \Big\|\partial_s\hat X_{t\to s}(\vx)-\teacher\big(\sg[\hat X_{t\to s}(\vx)],s\big)\Big\|^2,
 \label{eq:lmd}
\end{equation}
where $w$ is a distribution over time pairs and $\rho_t$ is a start-state distribution.
\citet{boffi2024flowmap,boffi2025howto} draw start states from the stochastic-interpolant marginal, spread start and target times over all of $[0,1]$, and parameterize $\hat X_{t\to s}(\vx)=\vx+(s-t)\,\vg_\theta(\vx,t,s)$ with a network $\vg_\theta$ conditioned on both times.
The time derivative $\partial_s\hat X$ then differentiates the network and is computed with the map by a JVP, at about $1.5$ times the cost of a forward pass~\citep{boffi2025howto}.

\paragraph{\method{} as a restricted Lagrangian map.}
For interval $k$ and $s\in[t_{k+1},t_k]$, let $\tau=(s-t_k)/\Delta_k$.
Since $(s-t_k)(\tau-1)=\Delta_k\tau(\tau-1)$, the curve in \eqref{eq:path} is the flow-map ansatz
\begin{equation}
 \hat X^{\theta}_{t_k\to s}(\vx_k)
 =\vx_k+(s-t_k)\Big[\vh_\theta+\sum_{j<K}(\tau-1)P_j(2\tau-1)\,\vc_{\theta,j}\Big]
 =\vx_{\theta,k}(\tau),
 \label{eq:ptd-map}
\end{equation}
with $\vh_\theta$ and $\vc_\theta$ evaluated once at $(\vx_k,t_k)$.
The bracket is the average velocity from $t_k$ to $s$, which Flow Map Matching and MeanFlow obtain from a network conditioned on both times~\citep{boffi2024flowmap,geng2025meanflow}; in \method{} it is a polynomial in $\tau$ whose coefficients come from one evaluation at the teacher's single-time interface.
Let $w$ place the start time on a grid point $t_k$ and draw $s$ uniformly from $[t_{k+1},t_k]$, and let $\rho_{t_k}$ be the law of the detached rollout state $\vx_k$.
The ansatz in \eqref{eq:ptd-map} then has three exact properties.
\begin{enumerate}
 \item[(a)] $\partial_s\hat X^{\theta}_{t_k\to s}(\vx_k)=\vv_{\theta,k}(\tau)$, a fixed linear combination of the outputs.
 \item[(b)] At $(t_k,s_i)$, the integrand of \eqref{eq:lmd} equals the $i$-th squared residual in \eqref{eq:ls}, so the equal-weight stratified loss in \eqref{eq:ls} is an unbiased estimate of \eqref{eq:lmd} on interval $k$, in value and in semigradient.
 \item[(c)] $\hat X^{\theta}_{t_k\to t_{k+1}}(\vx_k)=\vx_k+\Delta_k\vh_\theta$ for every $\vc_\theta$, so the sampler in \eqref{eq:sample} composes these one-interval maps without evaluating $\vc_\theta$.
\end{enumerate}
For (a), $s=t_k+\tau\Delta_k$ gives $\partial_s=\Delta_k^{-1}\partial_\tau$, and differentiating \eqref{eq:path} yields \eqref{eq:tangent}.
For (b), substituting $s_i=s(\tau_i)$ and (a) into \eqref{eq:lmd} gives $\|\vv_{\theta,k}(\tau_i)-\teacher(\sg[\vx_{\theta,k}(\tau_i)],s_i)\|^2=\|\vh_\theta+\sum_jd_j(\tau_i)\vc_{\theta,j}-\vu_i\|^2$ by \eqref{eq:reads}.
One uniform draw from each half of $[0,1]$ with equal weights has expectation $\int_0^1$ of this integrand, and uniform $\tau$ is uniform $s$ on the interval.
The law of the draws does not depend on $\theta$, so the same expectation holds for the semigradient, taken with the query states and $\rho_{t_k}$ fixed.
For (c), the factor $\tau-1$ vanishes at $s=t_{k+1}$.

Relative to \eqref{eq:lmd}, \method{} thus (i) fixes start times to the grid and keeps target times within the current interval, (ii) draws start states from its own detached rollout, (iii) places the target-time dependence in a fixed polynomial basis, and (iv) estimates the target-time integral with the stratified pair $m{=}2$.
It keeps the stop-gradient placement of \eqref{eq:lmd}, which the original Lagrangian loss of \citet{boffi2024flowmap} does not use.
\citet{boffi2024flowmap} state that their minimizer result holds for any start-state process with a strictly positive density, and use the interpolant so that the map is learned where it is evaluated; \method{}'s rollout serves the same purpose, training each map on the states its sampler visits.
Physics Informed Distillation corresponds to a single interval starting at $t=1$, with a time-conditioned network and finite-difference derivatives, and therefore generates in one step~\citep{tee2024pid}.

\paragraph{What the restrictions buy.}
Restrictions (i) and (iii) let the network keep the teacher's input $(\vx_k,t_k)$: the target time enters only through the basis, which spans a single interval.
Restriction (iii) makes (a) analytic: all $m$ target times share one student pass and ordinary backpropagation, with no JVP.
Property (c) separates the endpoint output from the interior shape: since $b_j(1)=0$, the endpoint depends on $\vh_\theta$ alone, so the curvature head can be removed without changing any deployed map, whereas a two-time map keeps its target-time input at deployment.
The curve family in \eqref{eq:path} contains the exact teacher segment $s\mapsto\Phi_{t_k\to s}(\vx_k)$ if and only if that segment is a polynomial of degree at most $K{+}1$ in $s$, so \method{} controls the endpoint through the residual rather than through exact representability.
If the Lipschitz condition of Appendix~\ref{app:family} holds for every start state in the support of $\rho_{t_k}$, coupling the two one-interval pushforwards of $\rho_{t_k}$ through the shared start state and squaring \eqref{eq:endpoint-residual-bound} gives, with $\hat\Phi_k(\vx_k)=\vx_k+\Delta_k\vh_\theta(\vx_k,t_k)$,
\begin{equation}
 W_2^2\big(\hat\Phi_{k\#}\rho_{t_k},\,\Phi_{t_k\to t_{k+1}\#}\rho_{t_k}\big)
 \le\Delta_k^2e^{2|\Delta_k|L_{\mathrm T}}\,
 \mathbb E_{\vx_k\sim\rho_{t_k}}\!\int_0^1\|\vr(\tau)\|^2\,\mathrm d\tau,
 \label{eq:interval-w2}
\end{equation}
where the right-hand expectation is the expected value of \eqref{eq:ls} at the current outputs.
This is the interval-level analogue of the Lagrangian Wasserstein bound of \citet{boffi2024flowmap}, whose exponent uses a one-sided Lipschitz constant.

\paragraph{Scope of the identity.}
The correspondence is exact for the value of \eqref{eq:ls} and for its ordinary semigradient in \eqref{eq:joint-gradients}.
The implemented routing (Appendix~\ref{app:h3_routing}) keeps the loss value and the $\vh$-gradient, before H3's modality normalization, and omits $2\bar d_j\bar{\vr}$ from the $\vc$-gradient.
The $\vh$-component of the output gradient therefore equals that of the Lagrangian semigradient, while the curvature component follows the centered residuals.
H3's final interval replaces the residual by the teacher-marched target in \eqref{eq:h3-anchor}.

\paragraph{Terminal Velocity Matching.}
TVM uses the same time convention and writes the map as a displacement $\vf(\vx_t,t,s)=\Phi^{\mathrm{data}}_{t\to s}(\vx_t)-\vx_t$, where $\Phi^{\mathrm{data}}$ is the flow map of the marginal velocity $\vu^{\mathrm{data}}$ between data and prior~\citep{zhou2026tvm}.
Its terminal velocity condition, $\frac{\mathrm d}{\mathrm ds}\vf(\vx_t,t,s)=\vu^{\mathrm{data}}(\Phi^{\mathrm{data}}_{t\to s}(\vx_t),s)$, is \eqref{eq:lagrangian} for that field, written for the displacement.
TVM trains from scratch: it replaces the unknown velocity at the predicted terminal state by the model's own velocity, with stop-gradient on the state and EMA weights for the target, and adds a flow-matching loss on data, so its first term has the form of Lagrangian self-distillation~\citep{boffi2025howto}.
Per query, the residual in \eqref{eq:ls} is TVM's terminal-velocity residual with the self-generated target replaced by the frozen teacher and $(t,s)$ restricted to $(t_k,s(\tau))$.
With $\vf_\theta=(s-t)\bm F_\theta(\vx_t,t,s)$, TVM computes $\frac{\mathrm d}{\mathrm ds}\vf_\theta=\bm F_\theta+(s-t)\partial_s\bm F_\theta$ with a JVP and backpropagates through it, using a fused attention kernel that supports the JVP backward pass.
In \method{}, the same terminal derivative is $\vv_{\theta,k}(\tau)$, read from the outputs through known basis derivatives.
TVM bounds $W_2^2$ between its one-step model distribution and the data by a weighted integral of its objective plus a constant, assuming the model velocity is Lipschitz, an assumption that motivates its architectural changes to the diffusion transformer.
\Eqref{eq:interval-w2} is the per-interval analogue for a frozen teacher: its residual is evaluated on the student's own curve, where \eqref{eq:ls} queries, its Lipschitz constant is the teacher's rather than the trained network's, and its reference distribution is the teacher's one-interval pushforward of $\rho_{t_k}$.

\paragraph{PDD.}
PDD predicts 128 interval-velocity outputs, or slots, on a fixed time grid, trains selected outputs against teacher midpoint Runge--Kutta block targets with block sizes 16--64, and at inference fuses four blocks of 32 slots into one endpoint projection per block~\citep{anonymous2026pdd}.
For both methods, the ideal target for a supervised interval is the teacher's exact mean velocity over that interval from its start state; for \method{} on interval $k$ this is $\vh_k^\star=(\Phi_{t_k\to t_{k+1}}(\vx_k)-\vx_k)/\Delta_k$.
The two methods approach this target differently.
A midpoint target queries the teacher at stage states placed by the teacher's own solver steps; it is fixed for a given start state and differs from the exact mean velocity by the solver's truncation error, to which PDD's trajectory-recovery result is subject (Appendix~\ref{app:related}).
\method{}'s targets are exact teacher velocities at states on the student's current curve, so they move with the student; the residual in \eqref{eq:ode-residual} measures the remaining mismatch and bounds the endpoint error through \eqref{eq:endpoint-residual-bound}.
When the student curve equals the teacher segment, $\vx_{\theta,k}(\tau)=\Phi_{t_k\to s(\tau)}(\vx_k)$, every query lies on the teacher path, the residual vanishes, and \eqref{eq:endpoint} gives $\vh_\theta=\vh_k^\star$, the quantity that a midpoint target approximates up to its truncation error.
H3's final interval is the one place where \method{} uses a solver target, the eight-step Euler march in \eqref{eq:h3-anchor}.

\paragraph{$\pi$-Flow.}
In $\pi$-Flow, one student evaluation at the interval start outputs a network-free policy $\pi(\vx,s)$, and the imitation loss matches $\pi$ to the teacher at states on a detached, numerically integrated rollout of the policy's own ODE~\citep{chen2026piflow}.
\method{}'s tangent defines such a policy, $\pi_k(\vx,s)=\vv_{\theta,k}\big((s-t_k)/\Delta_k\big)$, which depends on time only.
Its rollout from $\vx_k$ integrates in closed form to $\vx_{\theta,k}(\tau)$, so the imitation residual at a rollout state is $\vv_{\theta,k}(\tau_i)-\vu_i$, the residual in \eqref{eq:ls}.
For this policy class, on-policy imitation and \eqref{eq:lmd} therefore coincide.
For state-dependent policies, such as $\pi$-Flow's DX and GMFlow policies, the two losses have the same value but different gradients: imitation detaches the policy's state input, while the semigradient of \eqref{eq:lmd} differentiates through it.
Both methods thus match teacher velocities along the student's own trajectory; they differ in what is deployed.
$\pi$-Flow expands the output layer to emit its policy and integrates the policy with substeps at inference; \method{} deploys only the policy's interval average $\vh_\theta$, so each step is the update in \eqref{eq:sample} at the backbone's original output width.

\section{Implementation details}
\label{app:hparams}

\subsection{Time grids}
\label{app:grid}
We place student hop boundaries on the shifted grid $t=\frac{\gamma u}{1+(\gamma-1)u}$ with shift $\gamma{=}6$ and base time $u\in[0,1]$~\citep{esser2024sd3}.
The four hops are the images of the quarters $u\in\{1,\tfrac34,\tfrac12,\tfrac14,0\}$, giving $t=(0.999,\,0.947,\,0.857,\,0.667,\,0)$ (the first value is clamped) and $\Delta=(-0.052,\,-0.090,\,-0.190,\,-0.667)$. %
For H3, the student uses the release coupling $t^{\mathrm{aud}}=0.25t/(1-0.75t)$ for audio.
The LightX2V Turbo baseline retains its released video shift-6 and audio shift-3 sampler settings.
On Wan2.1-14B, both students use four blocks of 32 slots from the same 128-slot shift-6 grid.

\subsection{Training configurations}
\par\noindent\begin{minipage}{\linewidth}
  \centering\small
  \setlength{\tabcolsep}{4pt}
  \renewcommand{\arraystretch}{1.13}
  \captionof{table}{\textbf{Configurations of the evaluated PTD checkpoints.}
  The two backbones use different adaptation and final-hop training recipes; CFG denotes classifier-free guidance~\citep{ho2022cfg} and SLG skip-layer guidance.}
  \label{tab:hparams}
  \begin{tabular*}{\linewidth}{@{\extracolsep{\fill}}lll@{}}
    \toprule
     & Wan2.1-14B & MiniMax-H3 (33B) \\
    \midrule
    Height $\times$ width; frames & $480{\times}832$; 81 & $768{\times}1344$; 124 \\
    Frame rate & 16 fps & 24 fps \\
    Backbone training & Full fine-tuning & Per-hop LoRA, rank 128 \\
    Velocity projection & Dense, trainable & Per-hop LoRA \\
    Curvature terms $K$ & 8 & 8 \\
    Student-curve queries & 2 on every hop & 2 on hops 0--2 \\
    Final-hop supervision & Routed velocity LS & 8-step teacher march \\
    Teacher guidance & CFG 5; SLG, layer 12 & CFG-distilled release \\
    Learning rate (constant) & $10^{-5}$ & $10^{-4}$ \\
    Global batch & 256 & 512 \\
    Evaluated iteration & 64 & 150 \\
    EMA; warmup & None & None \\
    \bottomrule
  \end{tabular*}
\end{minipage}\par\vspace{8pt}

\paragraph{Prompt corpora and checkpoints.}
H3 uses 55{,}879 prompts drawn from ViMix, OpenVid, and VidProM and rewritten with the official H3 rewriter into its three-field audio-video template.
Text conditions are pre-encoded.
Our leakage filter removes the ViMix validation captions, the source of our evaluation prompts, from this corpus.
Training consumes $150\times512=76{,}800$ hop-level samples; since one rollout spans four iterations, these come from about $19{,}200$ prompt rollouts, approximately $0.34$ corpus passes.
Both Wan students train on randomly sampled prompts from the same ViMix prompt pool.
All training uses prompts without paired target videos.

\paragraph{MiniMax-H3 adaptation.}
The teacher is the frozen CFG-distilled H3 release; one teacher query requires one conditional forward pass.
The student shares this base and adds a rank-128 LoRA set per hop~\citep{hu2022lora}, including hop-specific LoRA on the original velocity projection.
No time-embedding or adaptive-layer-normalization LoRA is used.
The curvature head has eight Legendre terms.
Hops 0--2 use two stratified queries, with one uniform sample in each half-interval, and the gradient routing in Appendix~\ref{app:h3_routing}.
The last hop uses only the teacher-marched mean-velocity target in \eqref{eq:h3-anchor}.
Each modality's loss is divided by its detached value, with audio weight 1.
The learning rate is $10^{-4}$ for adapters and heads, with gradient clipping at 1, no EMA or warmup, and fp32 carried states.
The global batch is accumulated over 64 GPUs with one sample per GPU and eight accumulation rounds.
The reported checkpoint is iteration 150; its output has 124 frames and stereo 32-kHz audio.

\paragraph{Wan2.1-14B adaptation and baseline.}
We reproduce PDD with the training hyperparameters reported by \citet{anonymous2026pdd} and train PTD under the same training setting.
Both students train on Wan2.1-14B with global batch 256 and constant learning rate $10^{-5}$, without warmup, weight decay, or EMA, using bf16 computation and fp32 master weights; PTD accumulates this batch over 128 GPUs with two rounds and PDD over 64 GPUs with four.
Both methods query the same frozen teacher with CFG 5, skip-layer guidance (SLG) on layer 12, and the official negative prompt.
PTD trains the full transformer and dense velocity projection, together with the eight-term curvature head; the run has no LoRA parameters.
Every hop uses two stratified student-curve queries with the split gradient routing of Appendix~\ref{app:h3_routing}, without teacher marching.
PDD uses 128 slot heads, midpoint Runge--Kutta block targets, and training block sizes 16--64; inference fuses four blocks of 32 slots.
PDD is trained for 256 iterations, and its checkpoints at iterations 64, 96, and 256 reach overall VBench scores of 83.29, 83.67, and 82.83 (Table~\ref{tab:wan_checkpoint_selection}).
PTD reaches 84.57 at iteration 64, outperforming PDD both at the same iteration count and at PDD's best checkpoint, iteration 96, which serves as the baseline in the Astra, human-preference, and VBench comparisons.
The Wan PTD run continues to iteration 240; iteration 64 is its best evaluated checkpoint by overall VBench score.
Our PDD reproduction scores below the VBench values reported by \citet{anonymous2026pdd}, which we attribute to differences in training prompt data and evaluation prompts (the published rows use Self Forcing prompts) and to VBench evaluation noise.

\paragraph{Rollout and update accounting.}
Workers process the same hop at each training iteration, carry detached student endpoints to the next hop, and restart from noise after the fourth hop.
A complete rollout therefore spans four reported iterations; each iteration is one optimizer update, reported as an update in the main text.

\subsection{H3 final-hop supervision}
\label{app:h3}
Starting from the detached carried state $\vx_3$ at $t=0.667$, the H3 recipe marches the teacher to the endpoint and trains only the mean-velocity prediction:
\begin{equation}
 \vh_{\mathrm m}=\frac{\vx_{\mathrm{end}}-\vx_3}{\Delta_3},
 \qquad
 \mathcal L_{\mathrm{H3},3}
 =\left\|\vh_\theta-\sg[\vh_{\mathrm m}]\right\|^2.
 \label{eq:h3-anchor}
\end{equation}
There is no student-curve velocity loss or curvature-head supervision on this hop.
The eight Euler intervals use the shift-12 nodes
$0.667,0.632,0.591,0.541,0.481,0.405,0.308,0.180,0$.
The release's modality coupling determines the corresponding audio updates; the displayed target is per modality, with its own time interval and loss units.
The endpoint depends on the carried state and teacher integration, rather than the current hop's predicted curve.
This integration is used only during training.
Hops 0--2 retain the two-query student-curve loss.
In H3, we march only the final hop.
On high-noise hops the teacher trajectory bends sharply, and a marched endpoint folds this curvature into a target that is too hard for the student to learn in one evaluation; marching every hop softens local detail on Wan2.1-1.3B (Section~\ref{sec:ablations}).
The final hop is smoother: with $K{=}2$ Legendre terms, offline fits explain $90\%$ of its velocity variation, versus $46\%$ on the hop nearest noise (Figure~\ref{fig:curve_probe}).
We empirically found that marching this hop alone improves H3 visual detail at approximately 20\% higher training compute than the all-two-query recipe (Section~\ref{sec:h3}).

\paragraph{Model-specific recipes.}
The H3 checkpoint adds final-hop marching to the student-curve objective, whereas the Wan checkpoint uses the student-curve objective alone on all four hops.

\paragraph{Modality units.}
H3 follows the release's coupled audio/video time parameterization and modality-specific update rules.
Each modality loss is normalized by its detached value before their equally weighted combination.
Wan has a single modality and does not use this normalization.

\subsection{Removing the training scaffold}
\label{app:deployment}
The curvature head is unnecessary for the endpoint in \eqref{eq:curve-endpoint} and is removed at inference.
On H3, a hop's adapted projection, including the mean-velocity output, can be materialized as
\begin{equation}
 W^{(k)}=W+B^{(k)}A^{(k)},
\end{equation}
with adapter scale 1.
The low-rank product has the original projection's shape, so merging preserves the original layer types and output dimensions.
The shared frozen-base representation with active hop adapters computes the same linear maps in exact arithmetic, with different storage requirements.
On Wan, the fully fine-tuned backbone and dense mean-velocity projection are retained directly; there are no adapters to merge.
Both samplers take four steps using $\vx_{k+1}=\vx_k+\Delta_k\vh_\theta(\vx_k,t_k)$, without evaluating curvature coefficients.
The deployed models therefore retain their original architecture and latent output interface.

\section{Evaluation protocol and additional results}
\label{app:eval}

\paragraph{Models and scoring dimensions.}
The primary comparisons are \method{} at 150 updates against the released LightX2V Turbo v1.1 four-step LoRA on MiniMax-H3, and \method{} at 64 updates against our PDD reproduction at 96 updates on Wan2.1-14B.
All students use four network evaluations and seeds 0, 1, and 2 for each prompt.
Turbo retains its released sampler settings and LoRA scale 1.
GPT-6 Astra judges the outputs with medium reasoning effort and the V5 rubric reproduced in Appendix~\ref{app:gpt_astra_prompt}.
The four dimensions are dynamic quality (motion realism), static quality (individual-frame quality), naturalness, and diversity across the three seeds.
Greater motion alone does not improve the dynamic-quality score.
Audio is not evaluated.

\paragraph{Evaluation prompt construction.}
The 100-prompt evaluation set comprises two equally weighted groups, selected without judging model outputs.
V50 contains 50 held-out ViMix-14M~\citep{yang2025vimix} captions of real videos.
We shuffle the 804 eligible validation captions with seed 20260922, discard captions shorter than 15 words, and expand candidates using the public H3 prompt rewriter.
In that order, we retain the first 50 rewrites that pass the official-template check and begin with a live-action tag; accepted rewrites are used verbatim.
Our leakage filter excludes these source captions from the prompts used to train our H3 student.
D50 contains 50 Claude-generated, medium-dynamics, single-shot action prompts, with five prompts per category across ten categories.
These prompts were fixed before generation; they describe one subject performing a continuous physical action, with natural lighting and secondary motion, without cuts or stylization.
No prompts were added, removed, or edited after judging.
For H3, the judge receives the full official template prompt.
For Wan, we retain the video-description field and remove template headers, audio fields, and shot tags; the resulting prompts have median length 142 words.
Each model pair yields 600 clips and 400 judge calls over the 100 prompts.

\paragraph{Frames, blinding, and repetition.}
Each call receives 12 timestamped images, sampled uniformly over the clip including its first and last frame, in temporal order.
Each image is a $2\times3$ mosaic: one model per row and seeds 0, 1, and 2 across the columns.
The row labels show only position and seed, concealing model identities from the judge.
For H3, timestamps are $0.00$, $0.46$, $0.92$, $1.42$, $1.88$, $2.33$, $2.79$, $3.25$, $3.71$, $4.21$, $4.67$, and $5.12$ seconds; each video cell is resized to $256\times448$ pixels (height $\times$ width).
The Wan evaluation uses the same sampling protocol on its 16-fps clips.
Each row ordering is judged twice, giving four calls per prompt, and rows are mapped back to model identities before aggregation.
Output follows a JSON schema with keys \texttt{evidence}, \texttt{top\_row}, \texttt{bottom\_row}, and \texttt{preference}; evidence identifies relevant frames and seeds.

\paragraph{Score aggregation.}
Each call assigns one integer score from 1 to 4 per model and dimension, judging the three clips jointly.
We average the four scores within each prompt, then average across prompts with equal weights.
We compute paired score differences within prompts and bootstrap prompts to obtain 95\% confidence intervals; the H3 summaries use 4{,}000 percentile-bootstrap resamples with random seed 0.
The bootstrap unit is the prompt, with its seeds, time points, and repeated calls grouped together.
Reported differences use unrounded per-prompt scores and can differ slightly from subtracting the rounded model means.
We report each scoring dimension separately, aggregating scores across both row orders.
The automated ratings use the sampled video frames; human preference is evaluated through full-video playback.

\paragraph{Main-figure aggregation.}
For the four automated rows in Figure~\ref{fig:astra_main}, let $d_p$ be the mean of the four paired score differences for prompt $p$.
A \method{} win has $d_p\geq0.25$, a loss has $d_p\leq-0.25$, and a tie has $|d_p|<0.25$.
Writing these prompt counts as $W$, $T$, and $L$, the figure assigns half a credit to each method for every tie: \method{} occupies $(W+T/2)/N$ of the bar and the baseline occupies $(L+T/2)/N$, where $N=100$.
These shares are derived from prompt-level scores, not the judge's direct preference labels.
Table~\ref{tab:h3_evaluation} preserves the mean scores, paired confidence intervals, and original win/tie/loss counts.

\begin{table}[!ht]
  \centering
  \caption{\textbf{GPT-6 Astra scores on 100 evaluation prompts.} Mean scores on a 1--4 scale; $\delta$ is the paired mean difference (\method{} minus baseline), with 95\% bootstrap confidence intervals over 100 prompts. Each prompt has three seeds per model and four judge calls. W/T/L gives score-derived wins, ties, and losses from \method{}'s perspective.}
  \label{tab:h3_evaluation}
  \small
  \setlength{\tabcolsep}{3pt}
  \renewcommand{\arraystretch}{1.13}
  \begin{tabular*}{\linewidth}{@{\extracolsep{\fill}}lrrrrr@{}}
    \toprule
    Dimension & \method{} & Baseline & $\delta$ & 95\% CI & W/T/L \\
    \midrule
    \multicolumn{6}{@{}l}{\textit{MiniMax-H3: \method{} at 150 updates vs.\ released LightX2V Turbo v1.1}} \\
    Dynamic quality & 2.70 & 2.61 & $+0.09$ & $[-0.04,\,+0.21]$ & 45/27/28 \\
    Static quality  & 3.04 & 3.06 & $-0.03$ & $[-0.08,\,+0.02]$ & 6/86/8 \\
    Naturalness     & 3.09 & 2.98 & $+0.11$ & $[+0.06,\,+0.17]$ & 28/68/4 \\
    Diversity       & 2.81 & 1.55 & $+1.26$ & $[+1.15,\,+1.36]$ & 93/6/1 \\
    \midrule
    \multicolumn{6}{@{}l}{\textit{Wan2.1-14B: \method{} at 64 updates vs.\ our PDD reproduction at 96 updates}} \\
    Dynamic quality & 2.43 & 2.16 & $+0.27$ & $[+0.17,\,+0.36]$ & 47/41/12 \\
    Static quality  & 2.85 & 2.84 & $+0.01$ & $[-0.06,\,+0.07]$ & 12/78/10 \\
    Naturalness     & 2.82 & 2.60 & $+0.22$ & $[+0.13,\,+0.32]$ & 44/44/12 \\
    Diversity       & 2.47 & 2.52 & $-0.04$ & $[-0.17,\,+0.09]$ & 36/22/42 \\
    \bottomrule
  \end{tabular*}
  \par\vspace{3pt}
  \begin{minipage}{\linewidth}
    \scriptsize\raggedright
    Model means and paired differences are rounded independently; their displayed differences may disagree by 0.01. W/T/L is derived from the paired prompt-level score difference using a $0.25$ threshold.
  \end{minipage}
\end{table}

\paragraph{Results by prompt group.}
\label{app:astra_breakdown}
Table~\ref{tab:astra_supplement} separates the two prompt groups of the 100-prompt evaluation.
Every significant gain in Table~\ref{tab:h3_evaluation} holds within each prompt group: the paired 95\% intervals exclude zero in both V50 and D50 for H3 diversity ($+1.24$ and $+1.28$) and naturalness ($+0.19$ and $+0.04$), and for Wan dynamic quality ($+0.29$ and $+0.24$) and naturalness ($+0.19$ and $+0.25$).

\begin{table}[!ht]
  \centering
  \caption{\textbf{Astra results by prompt group.}
  Mean paired differences (\method{} minus baseline) with 95\% bootstrap confidence intervals over prompts.
  H3 uses LightX2V Turbo v1.1 as the baseline; Wan compares \method{} at 64 updates with our PDD reproduction at 96 updates.}
  \label{tab:astra_supplement}
  \small
  \setlength{\tabcolsep}{3pt}
  \renewcommand{\arraystretch}{1.23}
  \newcommand{\astrascore}[2]{\begin{tabular}[c]{@{}c@{}}$#1$\\[-1pt]{\scriptsize $#2$}\end{tabular}}
  \begin{tabular*}{\linewidth}{@{\extracolsep{\fill}}lcrrrr@{}}
    \toprule
    Prompt group & $n$ & \shortstack{Dynamic\\quality} & \shortstack{Static\\quality} & Naturalness & Diversity \\
    \midrule
    \multicolumn{6}{@{}l}{\textit{MiniMax-H3}} \\
    Live-action (V50) & 50 & \astrascore{+0.07}{[-0.09,+0.23]} & \astrascore{-0.06}{[-0.16,+0.04]} & \astrascore{+0.19}{[+0.09,+0.29]} & \astrascore{+1.24}{[+1.09,+1.39]} \\
    Action (D50) & 50 & \astrascore{+0.10}{[-0.10,+0.30]} & \astrascore{+0.00}{[+0.00,+0.00]} & \astrascore{+0.04}{[+0.01,+0.07]} & \astrascore{+1.28}{[+1.12,+1.43]} \\
    \midrule
    \multicolumn{6}{@{}l}{\textit{Wan2.1-14B}} \\
    Live-action (V50) & 50 & \astrascore{+0.29}{[+0.14,+0.45]} & \astrascore{+0.00}{[-0.10,+0.10]} & \astrascore{+0.19}{[+0.06,+0.33]} & \astrascore{-0.12}{[-0.30,+0.07]} \\
    Action (D50) & 50 & \astrascore{+0.24}{[+0.12,+0.38]} & \astrascore{+0.01}{[-0.06,+0.10]} & \astrascore{+0.25}{[+0.12,+0.38]} & \astrascore{+0.03}{[-0.15,+0.21]} \\
    \bottomrule
  \end{tabular*}
  \par\vspace{4pt}
  \begin{minipage}{\linewidth}
    \footnotesize\raggedright
    Positive differences favor \method{}.
    Dynamic quality rewards motion realism, not motion amount.
    All values, including zero-width intervals, retain the source precision.
  \end{minipage}
\end{table}

\paragraph{Wan2.1-14B VBench.}
\label{app:vbench}
We reproduce PDD~\citep{anonymous2026pdd} strictly following its reported training hyperparameters, with the same backbone, frozen teacher and guidance, ViMix training prompts, global batch size 256, and constant learning rate $10^{-5}$ as \method{} (Appendix~\ref{app:hparams}).
We train PDD for 256 updates and evaluate checkpoints at 64, 96, and 256 updates; update 96 has the highest overall score and serves as the PDD baseline for VBench, Astra, and the human study (Table~\ref{tab:wan_checkpoint_selection}).
\method{} reaches its best evaluated checkpoint after only 64 updates, where it already exceeds all three PDD checkpoints in both overall and quality scores; at the matched 64 updates, it leads PDD by $1.28$ overall points ($84.57$ vs.\ $83.29$).
Against the best evaluated PDD checkpoint, \method{} scores higher on 13 of the 16 VBench dimensions, with the largest gains in spatial relationship ($+8.73$), multiple objects ($+8.72$), and scene ($+5.67$) (Table~\ref{tab:wan_vbench_dimensions}).
VBench dynamic degree measures motion amount, whereas Astra's dynamic quality measures motion realism; \method{} significantly outperforms PDD in dynamic quality (Table~\ref{tab:h3_evaluation}).
As in Section~\ref{sec:protocol}, we report VBench as a reference; Astra ratings and blinded human votes are our primary evidence.
Our VBench runs use 946 prompts with five seeds and four network evaluations, whereas the published rows of Table~\ref{tab:main} use Self Forcing prompts~\citep{huang2025selfforcing}, except for rCM's proprietary prompts.

\paragraph{Embedding-based diversity.}
We report V-JEPA~2~\citep{assran2025vjepa2} and VideoMAE~V2~\citep{wang2023videomaev2} cosine-distance diversity scores on the five-seed VBench samples, following the metric convention of \citet{anonymous2026pdd}.
Larger distances indicate greater variation in the encoder's feature space.
These scores are separate from the VBench aggregates and from Astra's prompt-conditioned diversity rating on the 100 evaluation prompts.

\begin{table}[!ht]
  \centering
  \caption{\textbf{Wan2.1-14B checkpoints on VBench.} PDD is trained for 256 updates and evaluated at 64, 96, and 256 updates; update 96 has the highest overall score and is the best evaluated PDD checkpoint. \method{} is shown at its best evaluated checkpoint, 64 updates, where it exceeds all three PDD checkpoints in overall and quality scores. Bold marks the best overall score. All use four network evaluations.}
  \label{tab:wan_checkpoint_selection}
  \small
  \setlength{\tabcolsep}{5pt}
  \begin{tabular*}{\linewidth}{@{\extracolsep{\fill}}lrrrr@{}}
    \toprule
    Method & Updates & Overall & Quality & Semantic \\
    \midrule
    PDD & 64 & 83.29 & 84.43 & 78.71 \\
    PDD (best evaluated) & 96 & 83.67 & 85.24 & 77.40 \\
    PDD & 256 & 82.83 & 83.21 & 81.29 \\
    \midrule
    \method{} (best evaluated) & 64 & \textbf{84.57} & 85.44 & 81.10 \\
    \bottomrule
  \end{tabular*}
\end{table}

\begin{table}[!ht]
  \centering
  \caption{\textbf{Detailed VBench results for the best evaluated checkpoints.} PDD uses update 96 and \method{} update 64; both use four network evaluations. Higher is better under the VBench convention. Bold marks the higher score; $\delta$ is \method{} minus PDD in score points.}
  \label{tab:wan_vbench_dimensions}
  \small
  \setlength{\tabcolsep}{6pt}
  \renewcommand{\arraystretch}{1.08}
  \begin{tabular*}{\linewidth}{@{\extracolsep{\fill}}lrrr@{}}
    \toprule
    Dimension & PDD @96 & \method{} @64 & $\delta$ \\
    \midrule
    \multicolumn{4}{@{}l}{\textit{Quality dimensions}} \\
    Subject consistency & 94.45 & \textbf{95.79} & $+1.34$ \\
    Background consistency & 95.51 & \textbf{96.34} & $+0.83$ \\
    Temporal flickering & 98.74 & \textbf{98.80} & $+0.06$ \\
    Motion smoothness & \textbf{98.86} & 98.37 & $-0.49$ \\
    Dynamic degree & \textbf{85.83} & 76.11 & $-9.72$ \\
    Aesthetic quality & 63.45 & \textbf{66.73} & $+3.28$ \\
    Imaging quality & 66.75 & \textbf{68.40} & $+1.65$ \\
    \midrule
    \multicolumn{4}{@{}l}{\textit{Semantic dimensions}} \\
    Object class & 94.68 & \textbf{96.11} & $+1.43$ \\
    Multiple objects & 80.21 & \textbf{88.93} & $+8.72$ \\
    Human action & \textbf{97.60} & 97.40 & $-0.20$ \\
    Color & 82.87 & \textbf{86.81} & $+3.94$ \\
    Spatial relationship & 70.56 & \textbf{79.29} & $+8.73$ \\
    Scene & 49.10 & \textbf{54.77} & $+5.67$ \\
    Appearance style & 20.56 & \textbf{20.95} & $+0.39$ \\
    Temporal style & 24.28 & \textbf{24.61} & $+0.33$ \\
    Overall consistency & 26.35 & \textbf{26.88} & $+0.53$ \\
    \midrule
    Quality & 85.24 & \textbf{85.44} & $+0.20$ \\
    Semantic & 77.40 & \textbf{81.10} & $+3.70$ \\
    Overall & 83.67 & \textbf{84.57} & $+0.90$ \\
    \bottomrule
  \end{tabular*}
  \par\vspace{3pt}
  \begin{minipage}{\linewidth}
    \scriptsize\raggedright
    Differences are calculated from the displayed scores. Dynamic degree measures motion amount. Each checkpoint is its method's best evaluated checkpoint by overall VBench score.
  \end{minipage}
\end{table}

\paragraph{Human preference study.}\label{app:human}
Ten raters participated in blinded pairwise comparisons of \method{} with LightX2V Turbo v1.1 on MiniMax-H3 and with our PDD reproduction on Wan2.1-14B.
The final analysis contains 161 votes: 82 on H3 and 79 on Wan.
The checkpoints match the automated evaluation: \method{} at 150 updates on H3, and \method{} at 64 versus PDD at 96 updates on Wan; all generators use four network evaluations.
The study reuses the videos from the automated evaluation.
Each comparison contains 99 eligible prompts from that evaluation: 49 live-action captions and 50 action prompts, after one live-action prompt was excluded before data collection.
For each prompt, the interface presents two rows of three clips using matching seeds 0, 1, and 2, with model names and file identifiers concealed.
Raters choose the row they would rather watch, or a tie.
Picture quality, motion quality, naturalness, and diversity guide this single holistic preference.

Each session interleaves 20 items, comprising ten per model comparison and five from each prompt subset within each comparison.
The server prioritizes items with fewer votes and avoids assigning prompts already judged by the same rater.
Top/bottom placement is balanced per item with random tie-breaking, and presentation order is shuffled.
Clips play together and loop muted; an enlarged H3 clip can play audio.
The interface requests one full playback before voting, with a timeout fallback for browser playback events.
Raters may revisit an item; only the latest vote per session and item is retained.

Each retained vote contributes equally.
Let $W$, $T$, and $L$ denote \method{} wins, ties, and losses.
We summarize each comparison by the preference share $(W+T/2)/(W+T+L)$, splitting ties equally as in the automated bars of Figure~\ref{fig:astra_main}.
On H3, \method{} receives 45 wins, 14 ties, and 23 losses against LightX2V Turbo, a 63.4\% preference share.
On Wan, it receives 33 wins, 21 ties, and 25 losses against PDD, a 55.1\% preference share.
\method{} thus wins more votes than it loses in both comparisons.
Table~\ref{tab:human_preference} gives the action and live-action breakdowns together with non-tie win rates, Wilson 95\% confidence intervals, and two-sided exact sign-test $p$-values.
\begin{table}[!ht]
  \centering
  \caption{\textbf{Human preference results.}
  $n$ counts votes. Share is \method{}'s preference share with ties split equally, as in Figure~\ref{fig:astra_main}; the non-tie win rate excludes ties.
  We report Wilson 95\% confidence intervals for the non-tie win rate and two-sided exact sign-test $p$-values.}
  \label{tab:human_preference}
  \small
  \setlength{\tabcolsep}{3pt}
  \renewcommand{\arraystretch}{1.12}
  \begin{tabular*}{\linewidth}{@{\extracolsep{\fill}}lrrrrrrrr@{}}
    \toprule
    Subset & $n$ & \method{} wins & Ties & Losses & Share & \shortstack{Non-tie\\win rate} & 95\% CI & $p$ \\
    \midrule
    \multicolumn{9}{@{}l}{\textit{MiniMax-H3: \method{} at 150 updates vs.\ LightX2V Turbo v1.1}} \\
    All               & 82 & 45 & 14 & 23 & 63.4\% & 66\% & [54, 76] & 0.010 \\
    Action (D50)      & 43 & 22 & 10 & 11 & 62.8\% & 67\% & [50, 80] & 0.08  \\
    Live-action (V50) & 39 & 23 &  4 & 12 & 64.1\% & 66\% & [49, 79] & 0.09  \\
    \midrule
    \multicolumn{9}{@{}l}{\textit{Wan2.1-14B: \method{} at 64 updates vs.\ PDD at 96 updates}} \\
    All               & 79 & 33 & 21 & 25 & 55.1\% & 57\% & [44, 69] & 0.36 \\
    Action (D50)      & 40 & 21 &  8 & 11 & 62.5\% & 66\% & [48, 80] & 0.11 \\
    Live-action (V50) & 39 & 12 & 13 & 14 & 47.4\% & 46\% & [29, 65] & 0.85 \\
    \bottomrule
  \end{tabular*}
\end{table}

\section{Additional diagnostics}
\label{app:probes}

\subsection{Accuracy of two-query target estimation}
\label{app:read_ablation}
\paragraph{Protocol.}
The reported training recipe uses two stratified queries, one drawn independently and uniformly from each half of a hop.
We assess this sampling rule and a three-query alternative on a bank of 100 H3 teacher trajectories, stored at 64 integration steps on the shift-12 clock.
The diagnostic uses the training hop boundaries $[0.999,0.947,0.857,0.667,0]$ and interpolates the stored teacher velocities within each hop.
For each query count $m\in\{2,3\}$, we draw 32 independent stratified query sets per trajectory and form
\begin{equation}
 \bar{\vu}_m=\frac{1}{m}\sum_{r=1}^{m}\vu(\tau_r),\qquad
 \tau_r\overset{\mathrm{ind.}}{\sim}\mathcal U\!\left[\frac{r-1}{m},\frac{r}{m}\right).
\end{equation}
We compare this estimate with the reference mean velocity obtained from the stored endpoint displacement divided by the signed hop duration.
The squared discrepancy is normalized by the reference mean-velocity energy.
This is a teacher-path diagnostic of the query rule; training itself queries the student's predicted path.
We report video hops 0--2, the H3 hops trained with student-curve queries; the final hop is trained with the teacher-marched target instead (Appendix~\ref{app:h3}).

\paragraph{Two queries give a small error relative to the prediction residual.}
Table~\ref{tab:reads_quadrature} compares the normalized quadrature discrepancy with the mean-velocity error of the reported H3 checkpoint after 150 updates.
This student error is measured on 96 separate teacher-bank prompts, with the student started from the teacher-path state at each hop start, as the squared relative $L_2$ error of its predicted mean velocity with respect to the teacher's reference mean velocity on that hop.
The two-query discrepancy is $12.6\%$, $13.6\%$, and $13.3\%$ of this student error across the three hops, or approximately one eighth (Figure~\ref{fig:two_read_target}).
Three queries reduce the ratios to $6.6\%$, $6.1\%$, and $5.8\%$, respectively, at one additional teacher evaluation per supervised hop.
The training recipe therefore retains two queries.

\begin{table}[!ht]
  \centering
  \caption{\textbf{Two- and three-query estimation errors relative to student prediction error.}
  Video mean-velocity errors on H3 hops 0--2, normalized by the reference velocity energy.}
  \label{tab:reads_quadrature}
  \small
  \setlength{\tabcolsep}{5pt}
  \renewcommand{\arraystretch}{1.13}
  \begin{tabular*}{\linewidth}{@{\extracolsep{\fill}}crrrrr@{}}
    \toprule
    \multirow{2}{*}{Hop} & \multirow{2}{*}{Student MSE} & \multicolumn{2}{c}{Two queries} & \multicolumn{2}{c}{Three queries} \\
    \cmidrule(lr){3-4}\cmidrule(l){5-6}
    & & MSE & Relative & MSE & Relative \\
    \midrule
    0 & 0.3346 & 0.04218 & 12.6\% & 0.02224 & 6.6\% \\
    1 & 0.0114 & 0.00155 & 13.6\% & 0.00069 & 6.1\% \\
    2 & 0.0036 & 0.00048 & 13.3\% & 0.00021 & 5.8\% \\
    \bottomrule
  \end{tabular*}
  \par\vspace{4pt}
  \begin{minipage}{\linewidth}
    \footnotesize\raggedright
    All MSEs are normalized squared errors; relative values divide by the corresponding student MSE.
    Quadrature: 100 teacher trajectories, 32 random query sets per trajectory and query count.
    Student: reported H3 checkpoint (150 updates), squared teacher-anchored relative $L_2$ mean-velocity error on a separate 96-prompt bank.
    Relative values compare the quadrature and student error scales measured on their respective banks.
  \end{minipage}
\end{table}

\subsection{Within-hop trajectory learning on Wan2.1-14B}
\label{sec:trajectory_learning}
\paragraph{Protocol.}
We examine the learned trajectory shapes of PTD and our PDD reproduction on eight VBench prompts, using a shift-6 time grid and a 128-slot teacher reference path.
The comparison covers checkpoints at 32--224 student updates; PTD at 240 and PDD at 256 updates are listed separately as unmatched endpoints.
Checkpoints are matched by student-update count.
This diagnostic concerns latent trajectory shape and is separate from both VBench video scores and the 100-prompt Astra evaluation.

\paragraph{Bend magnitude and direction.}
Let $\mathbf b_S$ and $\mathbf b_T$ denote the student and teacher bend vectors measured by the probe in the middle half of each hop, $\tau\in[0.25,0.75]$.
For nonzero vectors, their magnitude ratio, directional alignment, and teacher-aligned projection are
\begin{equation}
  r=\frac{\|\mathbf b_S\|}{\|\mathbf b_T\|},\qquad
  \cos\theta=\frac{\langle\mathbf b_S,\mathbf b_T\rangle}
    {\|\mathbf b_S\|\,\|\mathbf b_T\|},\qquad
  q=\frac{\langle\mathbf b_S,\mathbf b_T\rangle}{\|\mathbf b_T\|^2}.
\end{equation}
The projection $q$ is an unbounded, signed coefficient that isolates the component of the student's bend parallel to the teacher's bend, normalized by teacher bend energy.
For a single vector pair, $q=r\cos\theta$; we average each metric independently over the probe samples.
Tables~\ref{tab:wan_curve_early} and~\ref{tab:wan_curve_late} give each statistic separately.

\paragraph{Calibration ratio.}
We also report $r/\cos\theta$.
For a vector pair with positive inner product, its reciprocal is the scalar $\alpha$ that minimizes $\|\alpha\mathbf b_S-\mathbf b_T\|^2$, giving the least-squares calibration of the predicted bend magnitude.

\paragraph{Trajectory shape across hops.}
The teacher bends most in the high-noise first hop: on the eight probe prompts, its velocity direction turns by $43.5^\circ$ within hop~0, compared with $4.7^\circ$ within the final hop.
In this strongly curved hop, PTD has a larger magnitude ratio $r$, direction cosine $\cos\theta$, and teacher-aligned projection $q$ than PDD at every shared checkpoint from 32 to 224 updates.
At 64 updates, $r$ is $0.081$ for PTD and $0.024$ for PDD, with cosines of $0.171$ and $0.139$ and projections of $0.014$ and $0.003$.
The best evaluated PTD checkpoint (64 updates) also exceeds the best evaluated PDD checkpoint (96 updates) on all three first-hop metrics ($r$: $0.081$ vs.\ $0.039$; $\cos\theta$: $0.171$ vs.\ $0.155$; $q$: $0.014$ vs.\ $0.006$).
On the smoother final hop, the teacher-aligned projections of the two methods differ by at most $0.005$ at every shared checkpoint; the final hop is also where the H3 recipe applies teacher marching (Section~\ref{sec:h3}).
Tables~\ref{tab:wan_curve_early} and~\ref{tab:wan_curve_late} report the full progression of magnitude, direction, and projection for all four hops.

\begin{table}[p]
  \centering
  \caption{\textbf{Wan2.1-14B trajectory-shape probe: Hops 0 and 1.} Summaries on eight VBench prompts. $r$: bend magnitude ratio; $\cos\theta$: bend-direction alignment; $q$: teacher-aligned bend projection. PTD and PDD share the listed update counts in the upper block; the final checkpoints in the lower block are unmatched.}
  \label{tab:wan_curve_early}
  \scriptsize
  \setlength{\tabcolsep}{2.5pt}
  \renewcommand{\arraystretch}{1.06}
  \begin{tabular*}{\linewidth}{@{\extracolsep{\fill}}rlrrrrrrrr@{}}
    \toprule
    & & \multicolumn{4}{c}{Hop 0} & \multicolumn{4}{c}{Hop 1} \\
    \cmidrule(lr){3-6}\cmidrule(l){7-10}
    Updates & Method & $r$ & $\cos\theta$ & $q$ & $r/\cos\theta$ & $r$ & $\cos\theta$ & $q$ & $r/\cos\theta$ \\
    \midrule
    32 & PTD & 0.047 & 0.200 & 0.010 & 0.234 & 0.030 & 0.020 & 0.001 & 1.485 \\
     & PDD & 0.017 & 0.139 & 0.002 & 0.123 & 0.025 & 0.023 & 0.001 & 1.065 \\
    \addlinespace[2pt]
    \textbf{64} & PTD & 0.081 & 0.171 & 0.014 & 0.476 & 0.046 & 0.028 & 0.001 & 1.668 \\
     & PDD & 0.024 & 0.139 & 0.003 & 0.172 & 0.025 & 0.023 & 0.001 & 1.090 \\
    \addlinespace[2pt]
    96 & PTD & 0.113 & 0.182 & 0.021 & 0.621 & 0.058 & 0.031 & 0.002 & 1.879 \\
     & PDD & 0.039 & 0.155 & 0.006 & 0.253 & 0.026 & 0.016 & 0.000 & 1.578 \\
    \addlinespace[2pt]
    128 & PTD & 0.138 & 0.226 & 0.033 & 0.613 & 0.059 & 0.035 & 0.002 & 1.673 \\
     & PDD & 0.054 & 0.160 & 0.009 & 0.339 & 0.027 & 0.026 & 0.001 & 1.008 \\
    \addlinespace[2pt]
    160 & PTD & 0.183 & 0.205 & 0.040 & 0.895 & 0.069 & 0.041 & 0.003 & 1.691 \\
     & PDD & 0.043 & 0.180 & 0.008 & 0.239 & 0.026 & 0.022 & 0.001 & 1.200 \\
    \addlinespace[2pt]
    192 & PTD & 0.185 & 0.241 & 0.046 & 0.768 & 0.071 & 0.035 & 0.003 & 2.028 \\
     & PDD & 0.053 & 0.208 & 0.011 & 0.257 & 0.035 & 0.020 & 0.001 & 1.723 \\
    \addlinespace[2pt]
    224 & PTD & 0.203 & 0.253 & 0.054 & 0.799 & 0.079 & 0.033 & 0.003 & 2.351 \\
     & PDD & 0.065 & 0.208 & 0.014 & 0.315 & 0.038 & 0.017 & 0.001 & 2.271 \\
    \midrule
    \multicolumn{10}{@{}l}{\textit{Unmatched final checkpoints}} \\
    240 & PTD & 0.226 & 0.300 & 0.072 & 0.751 & 0.075 & 0.033 & 0.003 & 2.261 \\
    256 & PDD & 0.075 & 0.235 & 0.018 & 0.321 & 0.039 & 0.023 & 0.001 & 1.710 \\
    \bottomrule
  \end{tabular*}
  \par\vspace{3pt}
  \begin{minipage}{\linewidth}
    \scriptsize\raggedright
    $r$ measures bend amplitude, $\cos\theta$ its direction, and $q$ its teacher-aligned projection. The sensitivity of $r/\cos\theta$ increases as $|\cos\theta|$ approaches zero. Bold: 64 updates, the best evaluated \method{} checkpoint; the best evaluated PDD checkpoint is at 96 updates.
  \end{minipage}
  \par\vspace{16pt}
  \normalsize
  \centering
  \caption{\textbf{Wan2.1-14B trajectory-shape probe: Hops 2 and 3.} Summaries on eight VBench prompts. $r$: bend magnitude ratio; $\cos\theta$: bend-direction alignment; $q$: teacher-aligned bend projection. PTD and PDD share the listed update counts in the upper block; the final checkpoints in the lower block are unmatched.}
  \label{tab:wan_curve_late}
  \scriptsize
  \setlength{\tabcolsep}{2.5pt}
  \renewcommand{\arraystretch}{1.06}
  \begin{tabular*}{\linewidth}{@{\extracolsep{\fill}}rlrrrrrrrr@{}}
    \toprule
    & & \multicolumn{4}{c}{Hop 2} & \multicolumn{4}{c}{Hop 3} \\
    \cmidrule(lr){3-6}\cmidrule(l){7-10}
    Updates & Method & $r$ & $\cos\theta$ & $q$ & $r/\cos\theta$ & $r$ & $\cos\theta$ & $q$ & $r/\cos\theta$ \\
    \midrule
    32 & PTD & 0.045 & 0.045 & 0.002 & 0.990 & 0.072 & 0.149 & 0.011 & 0.482 \\
     & PDD & 0.022 & 0.017 & 0.000 & 1.313 & 0.070 & 0.182 & 0.013 & 0.383 \\
    \addlinespace[2pt]
    \textbf{64} & PTD & 0.049 & 0.051 & 0.003 & 0.953 & 0.130 & 0.166 & 0.022 & 0.780 \\
     & PDD & 0.025 & 0.009 & 0.000 & 2.632 & 0.130 & 0.204 & 0.027 & 0.636 \\
    \addlinespace[2pt]
    96 & PTD & 0.050 & 0.053 & 0.003 & 0.953 & 0.168 & 0.178 & 0.031 & 0.941 \\
     & PDD & 0.033 & 0.019 & 0.001 & 1.700 & 0.160 & 0.214 & 0.035 & 0.748 \\
    \addlinespace[2pt]
    128 & PTD & 0.059 & 0.051 & 0.003 & 1.154 & 0.197 & 0.184 & 0.038 & 1.067 \\
     & PDD & 0.033 & 0.012 & 0.000 & 2.729 & 0.187 & 0.218 & 0.042 & 0.860 \\
    \addlinespace[2pt]
    160 & PTD & 0.064 & 0.052 & 0.004 & 1.216 & 0.217 & 0.197 & 0.045 & 1.102 \\
     & PDD & 0.039 & 0.011 & 0.000 & 3.431 & 0.208 & 0.222 & 0.047 & 0.934 \\
    \addlinespace[2pt]
    192 & PTD & 0.072 & 0.052 & 0.004 & 1.397 & 0.247 & 0.205 & 0.053 & 1.205 \\
     & PDD & 0.038 & 0.019 & 0.001 & 2.052 & 0.225 & 0.230 & 0.053 & 0.978 \\
    \addlinespace[2pt]
    224 & PTD & 0.072 & 0.051 & 0.004 & 1.420 & 0.277 & 0.209 & 0.061 & 1.325 \\
     & PDD & 0.035 & 0.019 & 0.001 & 1.882 & 0.244 & 0.234 & 0.058 & 1.043 \\
    \midrule
    \multicolumn{10}{@{}l}{\textit{Unmatched final checkpoints}} \\
    240 & PTD & 0.072 & 0.049 & 0.004 & 1.471 & 0.291 & 0.211 & 0.064 & 1.380 \\
    256 & PDD & 0.037 & 0.024 & 0.001 & 1.540 & 0.256 & 0.237 & 0.062 & 1.080 \\
    \bottomrule
  \end{tabular*}
  \par\vspace{3pt}
  \begin{minipage}{\linewidth}
    \scriptsize\raggedright
    $r$ measures bend amplitude, $\cos\theta$ its direction, and $q$ its teacher-aligned projection. The sensitivity of $r/\cos\theta$ increases as $|\cos\theta|$ approaches zero. Bold: 64 updates, the best evaluated \method{} checkpoint; the best evaluated PDD checkpoint is at 96 updates.
  \end{minipage}
\end{table}

\subsection{Curve expressiveness and frozen-feature prediction}
\label{app:curve_probes}
The diagnostics separate fitting a known teacher trajectory from predicting its velocity variation using a student's fixed features.
Figure~\ref{fig:curve_probe} summarizes the coefficient-count comparison; Tables~\ref{tab:curve_probe_basis} and~\ref{tab:curve_probe_sweep} give the basis and optimization details.

\paragraph{Offline trajectory fitting.}
Within each hop, we subtract the teacher's mean velocity and fit the remaining variation with $K$ derivative fields $d_j$.
Weighted least squares uses integration-step widths as weights and a double-precision QR solve.
We report the fraction of squared velocity-deviation energy explained by the fit, averaged over trajectories.
Thus, $99\%$ explained means a $1\%$ residual in velocity-deviation energy.
The H3 coefficient sweep uses eight 128-step video trajectories and the shift-6 hop grid; its values appear in Figure~\ref{fig:curve_probe}. %

\paragraph{Basis comparison.}
A separate H3 bank contains 1{,}000 trajectories integrated with 64 Euler steps on the native shift-12 clock.
Offline family comparisons use 20 fitting and 20 held-out trajectories; PCA modes are estimated only on the fitting set.
Only ten integration steps fall in the final hop, so the offline coefficient-count fits use the 128-step bank instead; the 1{,}000-trajectory bank serves the basis comparison and the frozen-feature probes.
Equivalent polynomial coordinates agree in offline fit to the reported precision.
The cubic B-spline and data-driven variants are listed separately because their spaces and coefficient counts differ from the $K{=}8$ polynomial comparison.

\paragraph{Frozen-feature prediction.}
We freeze an H3 student checkpoint trained for 150 updates with student-curve supervision on a shift-12 quarter grid.
Final-layer features are computed at the start of the final hop ($t=0.81$), using teacher states from the 1{,}000-trajectory bank.
A linear head predicts the teacher velocity deviation at three randomly sampled positions on each teacher path, with 500 prompts for fitting and 500 held out.
We compare a closed-form ridge fit with a zero-initialized head optimized by Adam $(0.9,0.99)$ for 40 batches of 250 prompts, with four optimizer steps per batch.
The reported explained fraction is
\begin{equation}
  E = 1-\frac{\|\widehat{\delta\vu}-\delta\vu\|^2}{\|\delta\vu\|^2}
    = 2r\cos\theta-r^2,
\end{equation}
where $r$ is the predicted-to-target amplitude ratio and $\theta$ is their angle at the sampled positions.
Across the tested $K\in\{2,3,4,8\}$ and every parameterization in Table~\ref{tab:curve_probe_basis}, final-hop scores remain near $45\%$ for ridge and $49$--$50\%$ for Adam at $10^{-3}$, so the Legendre default needs no basis-specific tuning at this learning rate.
At $10^{-4}$, coordinate choices produce larger finite-budget differences (Table~\ref{tab:curve_probe_basis}).
The final hop is also by far the most predictable: held-out scores on the earlier, higher-noise hops are $7$--$11\%$, $14$--$15\%$, and approximately $3.5\%$ for hops 0, 1, and 2, respectively. %
Together with the offline fits, in which $K{=}2$ already explains $90\%$ of final-hop variation versus $46\%$ near the noise end (Figure~\ref{fig:curve_probe}), these scores show that the final hop is smoother than the noise-end hop and more predictable than all earlier hops, which motivates applying teacher marching only on the final hop in the H3 recipe (Section~\ref{sec:h3}; Appendix~\ref{app:h3}).

\begin{table}[!ht]
  \centering
  \caption{\textbf{Basis choice: conditioning and finite-budget optimization.}
  H3 final-hop linear-probe scores (\%). Natural-scale and unit-RMS Legendre coordinates are distinguished.}
  \label{tab:curve_probe_basis}
  \small
  \setlength{\tabcolsep}{4pt}
  \renewcommand{\arraystretch}{1.13}
  \begin{tabular*}{\linewidth}{@{\extracolsep{\fill}}lrrrrr@{}}
    \toprule
    \multirow{2}{*}{Parameterization} & \multirow{2}{*}{$K$} & \multirow{2}{*}{$\kappa(G)\,\downarrow$} & \multirow{2}{*}{Ridge $\uparrow$} & \multicolumn{2}{c}{Adam $\uparrow$} \\
    \cmidrule(l){5-6}
     & & & & $10^{-3}$ & $10^{-4}$ \\
    \midrule
    \multicolumn{6}{@{}l}{\textit{Polynomial coordinates, $K=8$}} \\
    Legendre, natural scale & 8 & 7.9 & 45 & 50 & 46 \\
    Legendre, unit RMS & 8 & 6.3 & 45 & 50 & 48 \\
    Integrated Legendre, unit RMS & 8 & 1.05 & 45 & 49 & 48 \\
    Bernstein (B\'ezier), unit RMS & 8 & 792 & 45 & 50 & 49 \\
    \midrule
    \multicolumn{6}{@{}l}{\textit{Additional parameterizations; coefficient counts differ}} \\
    Clamped cubic B-spline & 4 & 8.4 & 45 & 50 & 49 \\
    Data modes (PCA) & 3 & -- & 45 & 50 & 47 \\
    \bottomrule
  \end{tabular*}
  \par\vspace{4pt}
  \begin{minipage}{\linewidth}
    \footnotesize\raggedright
    Adam scores are measured after the same 40-batch budget.
    $G$ is the population Gram matrix of the derivative fields.
    The additional parameterizations use the coefficient counts listed in the table.
    --: not measured.
  \end{minipage}
\end{table}

\paragraph{Population conditioning.}
Let $\widetilde{\mathbf d}(\tau)$ contain the $K$ derivative fields with the scaling used by a given probe.
We report $\kappa(G)$ for
\begin{equation}
  G = \mathbb E\!\left[\frac12\sum_{i=1}^{2}
    \widetilde{\mathbf d}(\tau_i)\widetilde{\mathbf d}(\tau_i)^{\mathsf T}\right],
\end{equation}
with one query position sampled uniformly from each half of the hop.
Natural-scale and unit-RMS coordinates are distinguished in Table~\ref{tab:curve_probe_basis}; the coefficient sweep uses unit-RMS fields.
This expected Gram matrix equals the uniform-query population Gram matrix.
Its condition number describes conditioning across the query distribution (Appendix~\ref{app:basis}). %

\begin{table}[!ht]
  \centering
  \caption{\textbf{Additional coefficient-count diagnostics.}
  Population conditioning and H3 final-hop frozen-feature prediction.}
  \label{tab:curve_probe_sweep}
  \small
  \setlength{\tabcolsep}{6pt}
  \renewcommand{\arraystretch}{1.14}
  \begin{tabular*}{\linewidth}{@{\extracolsep{\fill}}crr@{}}
    \toprule
    $K$ & \shortstack{Legendre, unit RMS\\$\kappa(G)\,\downarrow$} & \shortstack{H3 probe (\%) $\uparrow$\\Adam $10^{-4}$} \\
    \midrule
    1 & 1.0 & 39 \\
    2 & 1.0 & 40--42 \\
    3 & 1.6 & 44--48 \\
    4 & 2.2 & 45--49 \\
    \textbf{8} & \textbf{6.3} & \textbf{46--49} \\
    \bottomrule
  \end{tabular*}
  \par\vspace{4pt}
  \begin{minipage}{\linewidth}
    \footnotesize\raggedright
    Adam ranges are across the evaluated parameterizations after 40 batches.
    Bold marks the default $K$.
  \end{minipage}
\end{table}

\section{Extended related work}
\label{app:related}
We expand the comparisons in \Secref{sec:related}; Appendix~\ref{app:relations} gives a formal comparison with Lagrangian flow maps, TVM, PDD, and $\pi$-Flow.

\paragraph{Finite-interval mappings.}
Consistency models predict a common trajectory endpoint, while consistency trajectory models extend the mapping to pairs of times~\citep{song2023consistency,kim2024ctm}.
Flow Map Matching unifies consistency models, consistency trajectory models, and progressive distillation as learning the two-time flow map $\Phi_{s,t}$ of the underlying generative dynamics~\citep{boffi2024flowmap}.
Exact flow maps satisfy the semigroup property $\Phi_{u,t}\circ\Phi_{s,u}=\Phi_{s,t}$, so two consecutive jumps can be replaced by one~\citep{boffi2025howto}.
Progressive distillation trains one student step to match two teacher steps~\citep{salimans2022progressive}, and shortcut models condition on the step size and train one step to match two of their own consecutive half-size steps, a discretized semigroup condition~\citep{frans2025shortcut,boffi2025howto}.
Building on Flow Map Matching, \citet{boffi2025howto} convert distillation schemes into self-distillation and organize them into Eulerian, Lagrangian, and Progressive families; Align Your Flow scales continuous-time flow-map distillation~\citep{sabour2025ayf}.
MeanFlow parameterizes the flow map by the interval-average velocity and trains it with a differential identity whose target requires a Jacobian-vector product~\citep{geng2025meanflow}.
Its follow-ups recast this objective as instantaneous-velocity regression through an average-velocity network~\citep{geng2026imf}, analyze and rebalance its trajectory-matching and consistency components~\citep{alphaflow2025}, replace the identity with an algebraic interval-splitting consistency that needs no Jacobian-vector products~\citep{guo2025splitmeanflow}, convert pretrained flow models into flow maps~\citep{lee2026decoupled}, and use a restricted MeanFlow variant for stable distillation of large video models~\citep{zou2026disca}.
\method{}'s $\vh$ is the interval-average velocity learned by MeanFlow, but its targets come from the frozen teacher rather than from self-consistency or Jacobian-vector products of the student (Appendix~\ref{app:basis}).

\paragraph{Trajectory parameterizations and local supervision.}
Physics Informed Distillation parameterizes the trajectory as a time-conditioned network and approximates its derivative with numerical differences between student predictions~\citep{tee2024pid}.
DSNO learns continuous trajectories with Fourier temporal operators from solver-generated states and identifies a compact temporal spectrum in teacher trajectories~\citep{zheng2023dsno}.
$\pi$-Flow predicts a network-free velocity policy that imitates teacher velocities along detached policy rollouts and is integrated through substeps at inference~\citep{chen2026piflow}.
Transition Matching Distillation reuses backbone features while repeatedly evaluating a lightweight flow head, trained through distribution matching~\citep{nie2026tmd}.
PDD supervises selected interval outputs with teacher Runge--Kutta targets and fuses them into one endpoint projection per inference block; its trajectory-recovery result is subject to the Runge--Kutta approximation error~\citep{anonymous2026pdd}.
GENIE, B\'ezierFlow, and B\'ezier Distillation use higher-order or polynomial structure in solver derivatives, sampling schedulers, and interpolation paths, respectively~\citep{dockhorn2022genie,min2026bezierflow,feng2025bezier}.
\method{} instead predicts sample-dependent coefficients of an endpoint-preserving polynomial curve and deploys only its displacement output.

\paragraph{Distribution-level supervision.}
DMD and DMD2 match generated distributions through score-based supervision, and SiD optimizes a distributional discrepancy through score identities~\citep{yin2024dmd,yin2024dmd2,zhou2024sid}.
rCM regularizes consistency training with a score-based long-skip term, and Phased DMD applies distribution matching within subintervals; both report diversity gains relative to their DMD comparisons~\citep{zheng2025rcm,fan2025phaseddmd}.

\paragraph{Training states and released H3 accelerators.}
BOOT and FreeFlow study data-free distillation through bootstrapped trajectories and flow-map learning, respectively, and DMD2 and $\pi$-Flow also train on student-generated states~\citep{gu2024boot,tong2026freeflow,yin2024dmd2,chen2026piflow}.
For MiniMax-H3, FastH3 releases four-step DMD2 LoRA adapters, and LightX2V Turbo, our H3 baseline, provides four-step H3 adapters~\citep{fasth32026,modeltc2026h3turbo}.

\paragraph{Evaluating few-step video generators.}
On Wan2.1-14B, every published four-step student in Table~\ref{tab:main} scores above the 50-step teacher in overall VBench, and these students differ by sub-point margins, consistent with reports that VBench saturates on strong models~\citep{zheng2025vbench2,he2024videoscore,inbasekar2026worldjen}.
We therefore use blinded human votes and GPT-6 Astra judgments as primary evidence and report VBench for reference.

\clearpage
\section{Visual prompt for the first-page examples}
\label{app:teaser-prompts}
Figure~\ref{fig:teaser} compares two seeds for the snowboarding prompt below, using the 50-step MiniMax-H3 teacher, four-step LightX2V Turbo v1.1, and four-step \method{}. Each row shows five matching timestamps: 0.50, 1.71, 2.50, 3.42, and 4.75 seconds.
All displayed frames use the same center crop, retaining 80\% of the original width and height.

\subsection{Snowboarding}
\label{app:teaser-snowboard}
integrated\_multimodal\_description: [Shot 1] Live-action, cinematic, a low tracking shot follows a snowboarder in a dark grey shell jacket and a white helmet carving down a wide, freshly groomed slope under a clear sky at fast speed; ``PTD'' is printed in white block letters across the back of the jacket. The camera tracks forward at fast speed and shakes slightly, a few metres behind the board, as snow spray fans off every turn, the corduroy streaks past in a blur and the rider drops a gloved hand to brush the snow through a deep carve, the far ridgeline swinging across the frame. [Shot 2] At 00:02.800, the shot cuts to a wide static shot beside the slope as the rider lays the board over into a hard turn, a smooth arc of powder spraying up like a drawn curve, and the camera pans right with large amplitude at fast speed to follow the board straightening out down the fall line.

\clearpage
\section{GPT-6 Astra evaluation rubric}
\label{app:gpt_astra_prompt}
The automated evaluation uses \texttt{RUBRIC\_PAIR\_V5} from \texttt{astra\_judge.py}, selected by the default \texttt{-{}-rubric v5}, with GPT-6 Astra at \texttt{medium} reasoning effort.
Each call receives 12 chronologically ordered $2\times3$ mosaic frames, with one model per row and seeds 0/1/2 across the columns.
For each prompt and model pair, each of the two row orderings is judged twice, giving four calls in total.
The output is constrained by a JSON schema.
Below, \texttt{\{prompt\}} is replaced by the full generation prompt: the official audio-video template for H3 and the extracted video description for Wan.
The paper reports the four rubric scores and prompt-level outcomes derived from their differences; the preference labels requested below are not used.

\begingroup
\small\rmfamily\raggedright
\setlength{\parindent}{0pt}
\setlength{\parskip}{6pt}

You are a strict, objective judge comparing two text-to-video models on the same prompt. Each image is one time step (timestamp in the corner, in temporal order) of a 2x3 mosaic: the TOP row shows model A's three clips (seeds 0/1/2, left to right) and the BOTTOM row shows model B's three clips; the coloured bands only mark the row position and seed. Judge only visible evidence; the PROMPT is the target specification. Do NOT run any shell commands or tools.

For EACH row, score four metrics with an integer 1-4, judging the row as a whole (all three clips):

- dynamic\_quality: how REAL the movement itself looks, judged against how this action and camera move would look in genuine footage. Do NOT reward more motion: a larger or faster movement is not better; the right amount is whatever the action calls for. Assess four things: (a) natural kinematics: bodies, animals, vehicles and objects move with weight, momentum, balance and follow-through (gait cycles, lean into turns, recoil, settling), not floating, gliding, sliding or robotic; (b) realistic amplitude and tempo: the size and speed of the movement match the action and the physics of the scene; exaggerated, hyperactive or jerky motion is a defect, and so is damped, sluggish, half-frozen motion; (c) no stiffness: subjects should not hold a fixed pose while the camera or background moves, repeat the same pose across frames, or look puppet-like; secondary motion (hair, cloth, water, spray, dust, shadows) exists and follows the primary motion; (d) continuity of movement and camera: trajectories are smooth and continuous, the camera move is executed as one coherent path; no jitter, strobing, teleporting, sudden reversals, foot-skating, limbs or objects deforming while they move, or identity and appearance drifting over time; motion blur is consistent with the speed shown. Infer movement from how poses, positions and blur change between consecutive frames: plausible intermediate poses with consistent displacement = fluid; identical poses over a moving background = stiff; large unexplained jumps = broken. Judge the movement only, not image sharpness, detail or colour. 4 = movement indistinguishable from real footage on all four points; 3 = minor tells (a brief floaty or stiff moment, slight blur mismatch); 2 = clearly unnatural in places (puppet-like, exaggerated or damped, sliding, jerky camera); 1 = movement largely lifeless or broken (slideshow, chaotic deformation, impossible physics).

- static\_quality: image quality of the individual frames, judged as still pictures: clarity and fine detail, legible textures and text, correct anatomy and geometry, coherent structure, absence of in-frame artifacts (smearing, ghosting, extra or missing limbs, warped or disintegrating structure, blotches). Fine natural texture and film grain are not defects; over-sharpening and high contrast are not detail. Do not judge motion or temporal consistency here. 4 = no noticeable issues in any frame; 3 = minor; 2 = significant; 1 = severe.

- naturalness: how much the footage looks like real, professionally shot video (or, for stylised prompts, like a coherent production in the requested style): natural colour and tonal rendition (saturation, contrast, skin tones, highlight roll-off), believable materials, textures, lighting and depth, natural grain and motion blur, overall look and feel. Over-saturated, neon-tinted, over-sharpened, plastic or waxy "AI look" frames are defects, and so are washed-out, murky, flat or low-fidelity frames. 4 = indistinguishable from real footage / the requested style; 3 = minor tells; 2 = clearly synthetic; 1 = severe.

- diversity: how different the row's three clips are from each other in composition, subject appearance, camera and motion while all still fitting the prompt: 1 = near-identical, 2 = minor variations, 3 = clearly different in several aspects, 4 = distinct interpretations.

Then give a preference per axis (dynamic\_quality, static\_quality, naturalness, diversity) and overall: "top", "bottom" or "tie". Put timestamped evidence in "evidence" (name the specific frames/seeds that decide each axis). Return exactly one JSON object with keys evidence, top\_row, bottom\_row, preference.

PROMPT:\\
\{prompt\}
\par
\endgroup

\end{document}